%% file: main.tex
\documentclass[10pt,twocolumn,letterpaper]{article}
\usepackage[pagenumbers]{cvpr} 

\usepackage{graphicx}
\usepackage{amsmath}
\usepackage{amssymb}
\usepackage{booktabs}
\usepackage{bbm}
\usepackage{caption}
\usepackage{subcaption}
\usepackage{mathabx}
\usepackage{algorithm}
\usepackage{algpseudocode}
\usepackage{xcolor}
\usepackage{adjustbox}
\usepackage{braket}
\usepackage{comment}
\usepackage{pifont} 
\usepackage{multirow}
\usepackage{enumitem}
\usepackage{blindtext}
\usepackage[accsupp]{axessibility}

\usepackage[pagebackref,breaklinks,colorlinks]{hyperref}

\usepackage[capitalize]{cleveref}
\crefname{section}{Sec.}{Secs.}
\Crefname{section}{Section}{Sections}
\Crefname{table}{Table}{Tables}
\crefname{table}{Tab.}{Tabs.}

\def\confName{CVPR}
\def\confYear{2023}

\newcommand{\ourmethod}{\textsc{QuMF}\xspace} 
\newcommand{\ourmethoddec}{\textsc{DeQuMF}\xspace}
\newcommand{\ourmethodsa}{\textsc{QuMF (SA)}\xspace}
\newcommand{\ourmethodsadec}{\textsc{DeQuMF (SA)}\xspace} 
\newcommand{\ransac}{\textsc{RanSaC}\xspace}
\newcommand{\ransacov}{\textsc{RanSaCov}\xspace}
\DeclareMathOperator{\error}{error}
\newcommand{\suter}{\textsc{Hqc-RF}\xspace}

\graphicspath{ {./imgs/} }

\begin{document}

\title{Quantum Multi-Model Fitting} 

\author{Matteo Farina$^1$ \quad Luca Magri$^2$ \quad Willi Menapace$^1$\\ Elisa Ricci$^{1,3}$ \quad Vladislav Golyanik$^4$ \quad Federica Arrigoni$^2$ \\ 
\small
$^1$University of Trento \quad $^2$Politecnico di Milano \quad $^3$Fondazione Bruno Kessler \quad $^4$MPI for Informatics, SIC}

\maketitle

\begin{abstract} 
Geometric model fitting is a challenging but fundamental computer vision problem.
Recently, quantum optimization has been shown to enhance robust fitting for the case of a single model, while leaving the question of multi-model fitting open. 
In response to this challenge, this paper shows that the latter case can significantly benefit from quantum hardware and proposes the first quantum approach to multi-model fitting (MMF). 
We formulate MMF as a problem that can be efficiently sampled by modern adiabatic quantum computers without the relaxation of the objective function. 
We also propose an iterative and decomposed version of our method, which supports real-world-sized problems. 
The experimental evaluation demonstrates promising results on a variety of datasets.
The source code is available 
at: {\small\url{https://github.com/FarinaMatteo/qmmf}}.
\end{abstract}

\input{cready_sections/intro}
\input{cready_sections/background}
\input{cready_sections/method}
\input{cready_sections/related}
\input{cready_sections/experiments}
\input{cready_sections/limitations}
\input{cready_sections/conclusion}

{\small
\bibliographystyle{ieee_fullname}
\bibliography{bibliography}
}

\input{cready_sections/suppmat}
\end{document}

%% file: cready_sections/intro.tex
\section{Introduction}\label{sec:intro} 

Since the data volumes that AI technologies are required to process are continuously growing every year, the
pressure to devise more powerful hardware solutions is also increasing. To face this challenge, a promising direction currently pursued both in academic research labs and in leading companies is to exploit the potential
of quantum computing.
Such paradigm leverages quantum mechanical effects for computations and optimization, accelerating several important problems such as prime number factorization, database search \cite{Nielsen2010} and combinatorial optimization \cite{mcgeoch2014adiabatic}. %
The reason for such acceleration is that quantum computers leverage quantum parallelism of qubits, \textit{i.e.}, the property that a quantum system can be in a superposition of multiple (exponentially many) states and perform calculations simultaneously on all of them. 
\begin{figure}[t!]
    \centering 
    \includegraphics[width=\linewidth]{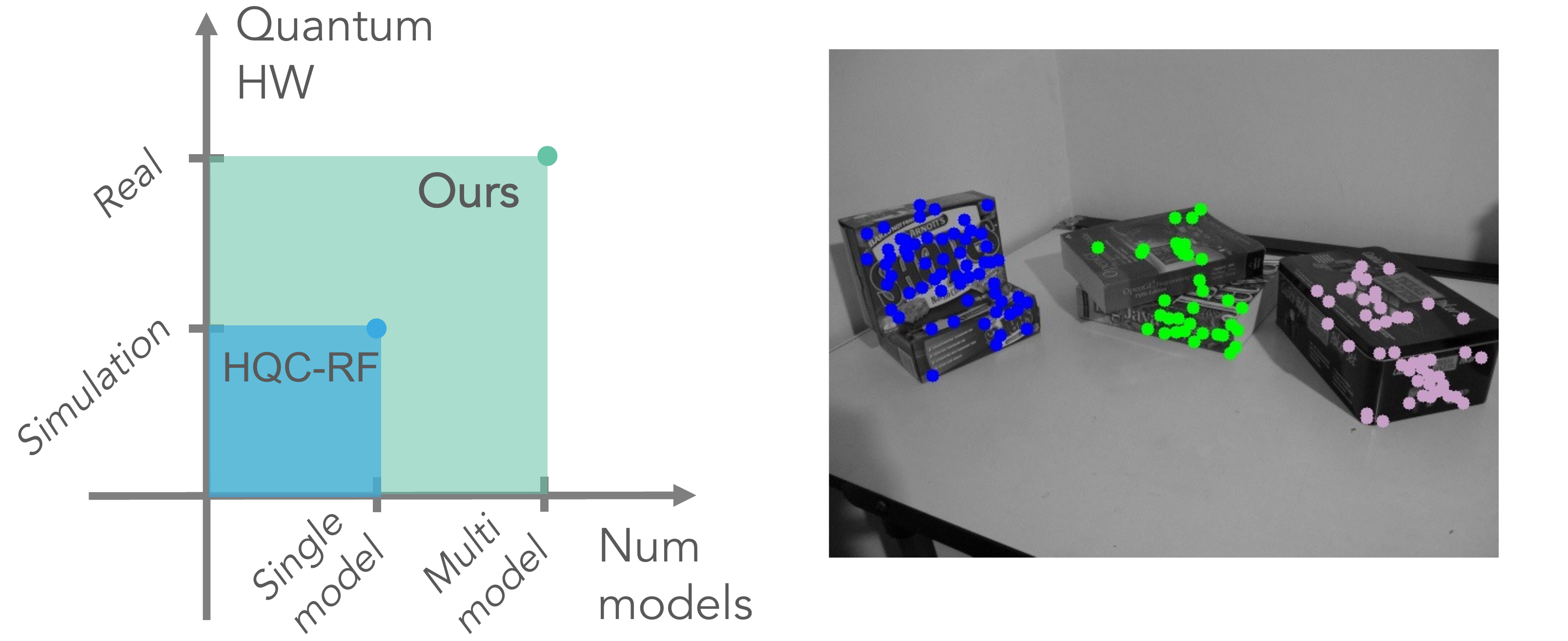} 
    \caption{
    \emph{Left:} Differences between our method and \suter \cite{Doan_2022_CVPR}. While \suter considers only a single model and is tested on quantum hardware with synthetic data, \ourmethod is also evaluated on real data on real quantum hardware. Although devised for multiple models, our method supports a single model likewise.  \emph{Right:} Qualitative results of \ourmethod on motion segmentation on the AdelaideRMF dataset \cite{wong2011dynamic}.
    \vspace{-2em}
    } 
 \label{fig_suter}
    \label{fig:teaser} 
\end{figure} 

Among the two quantum computing models, \textit{i.e.}, 
gate-based and Adiabatic Quantum Computers (AQCs), the latter recently gained attention in the computer vision community thanks to advances in experimental hardware realizations \cite{QuantumSync2021, SeelbachBenkner2021, Meli_2022_CVPR, Doan_2022_CVPR, Zaech_2022_CVPR, Yang_2022_CVPR, Arrigoni2022}. 
{At the present, AQCs provide sufficient resources in terms of the number of qubits, qubit connectivity and admissible problem sizes which they can tackle \cite{Boothby2020}, to be applied to a wide range of problems in computer vision.}
The AQC model is based on the adiabatic theorem of quantum mechanics \cite{BornFock1928} and designed for combinatorial problems (including $\mathcal{NP}$-hard) that are notoriously difficult to solve on classical hardware. 
Modern AQCs operate by optimizing objectives in the \emph{quadratic unconstrained binary optimization} (QUBO) form (see Sec.~\ref{sec:background}). However, many relevant tasks in computer vision cannot be trivially expressed in this form. 
Hence, currently, two prominent research questions in the field are: 1) \textit{Which problems in computer vision could benefit from an AQC?}, and 2) \textit{How can these problems be mapped to a QUBO form in order to use an AQC?} 

Several efforts have been recently undertaken to bring classical vision problems in this direction. 
Notable examples are works on graph matching \cite{SeelbachBenkner2020,SeelbachBenkner2021}, multi-image matching \cite{QuantumSync2021}, point-set registration \cite{golyanik2020quantum, Meli_2022_CVPR}, object detection \cite{LiGhosh2020},  multi-object tracking \cite{Zaech_2022_CVPR}, motion segmentation \cite{Arrigoni2022} and robust fitting \cite{Doan_2022_CVPR}.
Focusing on geometric model fitting, Doan \textit{et al.}~\cite{Doan_2022_CVPR} proposed an iterative consensus maximization approach to robustly fit a \emph{single} geometric model to noisy data. 
This work was the first to demonstrate the advantages of 
quantum hardware in robust single-model fitting with error bounds, which is an important and challenging problem with many applications in computer vision (\textit{e.g.,} template  recognition in a point set). 
The authors proposed to solve a series of linear programs on an AQC and demonstrated promising results on synthetic data. 
They also showed experiments for fundamental matrix estimation and point triangulation using simulated annealing (SA) \cite{Kirkpatrick1983}. 
SA is a classical global optimization approach, that, in contrast to AQCs, can optimize arbitrary objectives and is a frequent choice when evaluating quantum approaches (see Sec.~\ref{sec:background}). 
\suter \cite{Doan_2022_CVPR} takes advantage of the hypergraph formalism to robustly fit a single model. It is not straightforward to extend it to the scenario where \emph{multiple} models are required to explain the data.

Multi-model fitting (MMF) is a relevant problem in many applications, such as 3D reconstruction, where it is employed to fit multiple rigid moving objects to initialize multi-body Structure from Motion \cite{OzdenSchindlerAl10,ArrigoniRicciAl22}, or to produce intermediate interpretations of reconstructed 3D point clouds by fitting geometric primitives\cite{MagriLeveniAl21}. 
Other scenarios include face clustering, body-pose estimation, augmented reality and image stitching, to name a few.

This paper proposes \ourmethod, \textit{i.e.,} the first quantum 
MMF approach. 
We propose to leverage the advantages of AQCs in optimizing combinatorial QUBO objectives to explain the data with \emph{multiple} and \emph{disjoint} geometric models. 
Importantly, \ourmethod does not assume the number of disjoint models to be known in advance.
Note that the potential benefit from AQCs for MMF is higher than in the single-model case: when considering multiple models the search space scales exponentially with their number, making the combinatorial nature of the problem even more relevant.
Furthermore, we show that \ourmethod can be easily applied to single-model fitting even though not explicitly designed for this task. 
We perform an extensive experimental evaluation on quantum hardware with many large-scale real datasets and obtain competitive results with respect to %
both classical and quantum methods. 
Figure \ref{fig:teaser} depicts a visual comparison between \suter \cite{Doan_2022_CVPR} and \ourmethod. 

\smallskip
\textbf{Contributions.} In summary, the primary technical contributions of this paper are the following: 
\begin{itemize}[topsep=0.1em, itemsep=0.05em] 
    \item We bring multi-model fitting, a fundamental computer vision problem with combinatorial nature, into AQCs; 
    \item We introduce \ourmethod, demonstrating that it can be successfully used both for single and multiple models;
    \item We propose \ourmethoddec, a decomposition policy allowing our method to scale to large-scale problems, %
    overcoming the limitations of modern quantum hardware. 
\end{itemize}
The following section provides the background on AQCs and how to use them to solve QUBO problems. 
After introducing \ourmethod and \ourmethoddec in Sec.~\ref{sec:proposed-methodology}, we discuss related work in Sec.~\ref{sec:related-work}. Experiments are given in Sec.~\ref{sec:experimental-results}. Limitations and Conclusion are reported in Sec.~\ref{sec:limitations} and \ref{sec:conclusion}, respectively.

%% file: cready_sections/background.tex
\section{Background}\label{sec:background} 
The \textbf{QUBO formulation} in a \emph{binary} search space is an optimization problem of the following form:
\begin{equation}
    \min_{\mathbf{y} \in \mathbb{B}^d}{\mathbf{y}^{\mathsf{T}}Q\mathbf{y} + \mathbf{s}^{\mathsf{T}}\mathbf{y}}, 
    \label{eq_qubo}
\end{equation}
where $\mathbb{B}^d$ denotes the set of binary vectors of length $d$, $Q \in \mathbb{R}^{d \times d}$ is a real symmetric matrix and, in a QUBO with $d$ variables, their linear coefficients are packed into $\mathbf{s} \in \mathbb{R}^d$. When the problem is subject to linear constraints of the form $A \mathbf{y} = \mathbf{b} $, a common approach is to relax such constraints and reformulate the QUBO as:
\begin{equation}
    \min_{\mathbf{y} \in \mathbb{B}^d}{\mathbf{y}^{\mathsf{T}}Q\mathbf{y} + \mathbf{s}^{\mathsf{T}}\mathbf{y} + \lambda||A\mathbf{y} - \mathbf{b}||_{2}^{2}}, 
    \label{eq:qubo_soft}
\end{equation}
where $\lambda$ must be tuned to balance the contribution of the constraint term. Unravelling such a term leads to:
\begin{equation} \label{eq:soft-constrained-qubo}
    \min_{ \mathbf{y} \in B^d}{\mathbf{y}^{^{\mathsf{T}}}\widetilde{Q}\mathbf{y} + \Tilde{\mathbf{s}}^{^{\mathsf{T}}}\mathbf{y}}
\end{equation}
where the following simple substitutions are implied:
\begin{equation} \label{eq:general-constraints-subs}
    \widetilde{Q} = Q + \lambda A^{\mathsf{T}} A, \ \quad 
    \Tilde{\mathbf{s}} = \mathbf{s} - 2\lambda A^{\mathsf{T}} \mathbf{b}
\end{equation}
For the proof, the reader can refer to, e.g., Birdal \textit{et al.}~\cite{QuantumSync2021}. 

\smallskip
\textbf{Adiabatic Quantum Computers (AQCs)} are capable of solving QUBO problems. 
An AQC is organized as a fixed and architecture-dependent undirected graph, whose nodes correspond to \emph{physical qubits} and whose edges correspond to \emph{couplers} (defining the connectivity pattern among qubits) \cite{Dattani2019}. 
Such graph structure, in principle, can be mapped to a QUBO problem \eqref{eq_qubo} as follows: each physical qubit represents an element of the solution (i.e., of the binary vector $\mathbf{y}$), while each coupler models an element of $Q$ (i.e., the coupler between the physical qubits $i$ and $j$ maps to entry $Q_{ij}$). 
An additional weight term is assigned to each physical qubit, modelling the linear coefficients in $\mathbf{s}$.

Because AQC graphs are not fully connected, mapping an arbitrary QUBO to quantum hardware requires additional steps. 
This concept is easier to visualize if the notion of \emph{logical graph} is introduced: the QUBO itself can also be represented with a graph, having $d$ nodes (called \emph{logical qubits}\footnote{Hereafter we will sometimes omit the term ``logical'' or ``physical'' as it will be clear from the context which type of qubits are being involved.}), corresponding to the $d$ entries of $\mathbf{y}$ and edges corresponding to the non-zero entries of $Q$.

The logical graph undergoes \emph{minor embedding}, where an algorithm such as 
\cite{Cai2014} maps the logical graph to the physical one. During this process, a single logical qubit can be mapped to a set of physical qubits, \textit{i.e.,} a \emph{chain}. 
All the physical qubits in a chain must be constrained to share the same state during annealing 
\cite{pelofske2020advanced}. 
The magnitude of such constraint is called \emph{chain strength} and the number of physical qubits in the chain is called \emph{chain length}. 
An equivalent QUBO that can be directly embedded on the physical graph is obtained as the output of minor embedding. 

Then, optimization of the combinatorial objective takes place: we say that the AQC undergoes an \emph{annealing} process. In this phase, the physical system transitions from a high-energy initial state to a low-energy final state, representing the solution of the mapped QUBO problem, according to the laws of quantum mechanics \cite{Farhi2001,mcgeoch2014adiabatic}.
At the end of the annealing, a binary assignment is produced for each physical qubit.
A final postprocessing step is needed to go from physical qubits back to logical ones, thus obtaining a candidate solution for the original QUBO formulation (prior to minor embedding). 
Due to noise during the annealing process, the obtained solution may not correspond to the global optimum of the QUBO problem. 
In other terms, the annealing is inherently probabilistic and has to be repeated multiple times. 
The final solution is obtained as the one that achieves the minimum value of the objective function (lowest energy). 
Performing multiple annealings of the same QUBO form is called \emph{sampling}. 

It is common practice \cite{SeelbachBenkner2021, Doan_2022_CVPR, Zaech_2022_CVPR, Arrigoni2022} to test the QUBO objective on classic computers using \emph{simulated annealing} (SA) \cite{Kirkpatrick1983}, a probabilistic optimization technique able to operate in the binary space.
SA is probabilistic, so solutions are sampled multiple times, as it happens during quantum annealing. 
In contrast, SA operates on the original QUBO without embedding (i.e. directly on the logical graph) since no physical graph is involved. 
Importantly, SA can be used to evaluate performance on larger problem instances, in anticipation of larger AQC architectures.

%% file: cready_sections/method.tex
\section{The Proposed Method}
\label{sec:proposed-methodology}

In this section we derive our approach: first, we define an optimization function for the multi-model fitting task; then, we align it to the standard QUBO formulation, in order to enable quantum optimization; finally, we propose a decomposed version that can manage large-scale problems. 

\subsection{Problem formulation}

Consider a set of data points $ X= \{ x_1, \dots, x_n \} $ and a set $\Theta =\{ \theta_1, \dots, \theta_m \}$ of tentative and redundant (parametric) models, which, in practice, are obtained by a random sample procedure as done in \ransac \cite{ransac}.  Specifically, the models are instantiated by randomly selecting the minimum number of points necessary to unambiguously constrain the parameters of a model (\emph{e.g.}, 2 points for a line, 8 for a fundamental matrix). 
The aim of multi-model fitting is to extract from $\Theta$ the ``best'' models that describe all the data $X$. 
Equivalently, MMF can be seen as a clustering problem, where points belonging to the same model must be grouped together. 
This is an ill-posed task since different interpretations of the same data are possible and recovering the best models requires the definition of a proper notion of optimality. 
To this end, the problem has been cast as the optimization of an objective function encoding different regularization strategies to disambiguate between interpretations and include additional prior about the data. 
The choice of the objective function is critical and determines the performance of the algorithm. 

\subsection{Choice of Optimization Objective}
\label{sec:mmf-optimization}

In this paper, we focus on the objective function of \ransacov (Random Sample Coverage) \cite{magrifusiello16} where MMF is treated as a set-cover problem. 
This choice is due to: i) the \emph{simplicity} of the objective, which does not involve complex regularization terms as well as does not require to know in advance the number of models; ii) its \emph{combinatorial nature}, that is well capitalized by a quantum formulation; iii) its \emph{generality}, as this approach manages the broad situation of fitting {multiple} models, in contrast to previous work on quantum model fitting \cite{Doan_2022_CVPR}.
\ransacov \cite{magrifusiello16} casts MMF as an integer linear program. Such formulation is 
based on the notion of \textit{Preference-Consensus Matrix}, which is commonly denoted by $P$ and defined as:
\begin{equation}
    P[i,j]=
    \begin{cases}
        1 & \text{if $\error(x_i, \theta_j) < \epsilon$}\\
        0 & \text{otherwise}\\
    \end{cases}
    \label{eq:preference}
\end{equation}
where: $x_i$ and $\theta_j$ are the $i$-th point in the dataset and the $j$-th sampled model respectively; $\error$ expresses the residual of point $x_i$ w.r.t. model $\theta_j$ and $\epsilon$ is the inlier threshold to assign the point to the model. $P$ is a matrix of size is $n \times m$, where $n$ and $m$ are the number of points and models, respectively. Intuitively, the columns of $P$ represent the consensus sets of models, while the rows are points' preference sets. 

In this setup, selecting the subset of sampled  models that best describe the data, is equivalent to selecting the minimum number of columns of $P$ that explain all the rows. More formally:
\begin{equation} \label{eq:set-cover}
    \begin{gathered}
        \min_{\mathbf{z} \in \mathbb{B}^m}{\mathbbm{1}_{m}^{\mathsf{T}}\mathbf{z}} \quad 
        \text{s.t. } P\mathbf{z} \geq \mathbbm{1}_{n}
    \end{gathered}
\end{equation}
where $\mathbbm{1}$ denotes a vector of ones (whose length is given as subscript) and $ \mathbf{z}$ is a binary vector representing which models (i.e., columns of the preference matrix $P$) are selected.
Solving \eqref{eq:set-cover} entails minimizing the number of selected models, while ensuring that each point is explained by \emph{at least} one of them (as expressed by the inequality constraint). This problem is also known as \emph{set cover}: we are looking for a \emph{cover} of the data (i.e., a collection of sets whose union contains all the points) consisting of a minimum number of models, thus discouraging redundancy.

Our goal is to express MMF as a QUBO, in order to make it suitable for optimization on an AQC and capitalize all the advantages of quantum computing.
In general, posing a computer vision problem as a QUBO may not be straightforward, hence a careful choice of assumptions is %
essential.
First, in this paper, we assume that the true number of models/clusters (denoted by $k$) is \emph{unknown}, so our approach is %
general. 
Second, we assume that outliers are not present in the data, \emph{i.e.}, there exist \emph{exactly} $k$ models that describe all the points. 
This is not a major limitation because,  when outliers are present in the data, the models fitted to outliers could be easily recognized and discarded a-posteriori, using \emph{a-contrario} reasoning \cite{moulon2013global}, as usually done by MMF algorithms based on clustering \cite{MagriFusiello14,TepperSapiro14}. 
Third, we focus on the case where \emph{disjoint} models are being sought, which is highly common in practical scenarios. In this situation, the inequality in \eqref{eq:set-cover} can be turned into an equality constraint. 

These assumptions allow to rewrite the problem in terms of \textbf{disjoint set cover}, which is the basis of our work: 
\begin{equation} \label{eq:disjoint-set-cover}
    \begin{gathered}
        \min_{\mathbf{z} \in \mathbb{B}^m}{\mathbbm{1}_{m}^{\mathsf{T}}\mathbf{z}}\quad
        \text{s.t. } P\mathbf{z} = \mathbbm{1}_{n}.
    \end{gathered}
\end{equation}
The above constraint, instead of ensuring that each point is explained by \textit{at least} one selected model, forces each point to be explained by \textit{exactly} one model, hence imposing the consensus sets of the selected models are disjoint. Problem \eqref{eq:disjoint-set-cover} is the starting point of our QUBO formulation.

\subsection{QUBO Formulation}
\label{sec:mmf-qubo-formulation}

Problem \eqref{eq:disjoint-set-cover} is an integer linear program which is known to be NP-hard, hence it represents a suitable candidate to exploit the advantages of quantum optimization. To accomplish such a task, we need to turn  \eqref{eq:disjoint-set-cover} into a QUBO.
Being a linear program, Problem \eqref{eq:disjoint-set-cover} is, in particular, a special case of a quadratic program with linear equality constraints. Hence, if we pose constraints as soft ones (instead of hard ones), then we can rewrite \eqref{eq:disjoint-set-cover} as \eqref{eq:qubo_soft} using the following correspondences:
\begin{equation} 
\label{eq:mmf-qubo-terms}
Q=0, \ \quad \mathbf{s} = \mathbbm{1}_m, \  \quad A = P , \  \quad \mathbf{b} = \mathbbm{1}_n 
\end{equation}
where optimization is carried out over $\mathbf{z} \in \mathbb{B}^m$, that represents the choice of a subset of optimal models.
Putting together the equalities in \eqref{eq:general-constraints-subs} and \eqref{eq:mmf-qubo-terms}, it follows that:
\begin{equation} \label{eq:mmf-constrained-substitutions}
        \widetilde{Q} = \lambda P^{\mathsf{T}} P, \ \quad
        \Tilde{ \mathbf{s} } = \mathbbm{1}_m - 2\lambda P^{\mathsf{T}}\mathbbm{1}_n
\end{equation}
where no matrix of quadratic coefficients is present in $\widetilde{Q}$ as the considered formulation only contains linear terms, i.e., $Q=0$. 
Hence, Problem \eqref{eq:disjoint-set-cover} can be turned into a QUBO of the form \eqref{eq:soft-constrained-qubo}, namely:
\begin{equation} \label{eq:mmf-qubo}
    \min_{ \mathbf{z} \in \mathbb{B}^m}{\lambda \mathbf{z}^{\mathsf{T}} (P^{\mathsf{T}} P )\mathbf{z} + (\mathbbm{1}_m-2\lambda P^{\mathsf{T}}\mathbbm{1}_n)^{\mathsf{T}} \mathbf{z}} .
\end{equation}
Once a QUBO formulation has been obtained for a specific problem as in \eqref{eq:mmf-qubo}, an AQC can be leveraged to find an optimal solution, following the procedure from Sec.~\ref{sec:background}. 
This gives rise to our quantum approach for multi-model fitting, hereafter named \ourmethod.

\subsection{Iterative Decomposed Set-Cover}
\label{sec:id-qmmf}

\begin{algorithm}[t]
\caption{\ourmethoddec}\label{algo:decomposition}
\small 
\begin{algorithmic}
\Require $P$, $s$
\While{$|P.\operatorname{columns}| > s$}
\State $\text{subproblems} = \operatorname{ColumnPartition}(P, s)$
\State $i \gets 0$
\While{$i < |\text{subproblems}|$}
\State $\mathcal{J}_i \gets$ models in the $i$-th subproblem
\State $P_{\mathcal{J}_i}\gets $  $P$ retaining only the $\mathcal{J}_i$ columns
\State $\mathbf{z}_i = \ourmethod(P_{\mathcal{J}_i})$
\State remove from $P$ columns $\ P_{\mathcal{J}_i}[:,1-\mathbf{z}_i]$%
\State $i \gets i+1$
\EndWhile
\EndWhile
\State $\mathbf{z} \gets \ourmethod(P)$\\
\Return $P$[:,$\mathbf{z}$]
\end{algorithmic}
\end{algorithm}

While the capabilities of quantum hardware are rapidly evolving \cite{Boothby2020,Jurcevic2021}, current AQC architectures offer a limited number of physical qubits, thus the maximum dimension of the problems that can be solved is constrained in terms of $m$.
In order to overcome this limitation, we present a decomposed approach that iteratively prunes out columns of the preference matrix $P$ and reduces the dimension of the problem to fit the number of available physical qubits. 

\begin{figure}
    \centering
    \includegraphics[width=0.8\columnwidth]{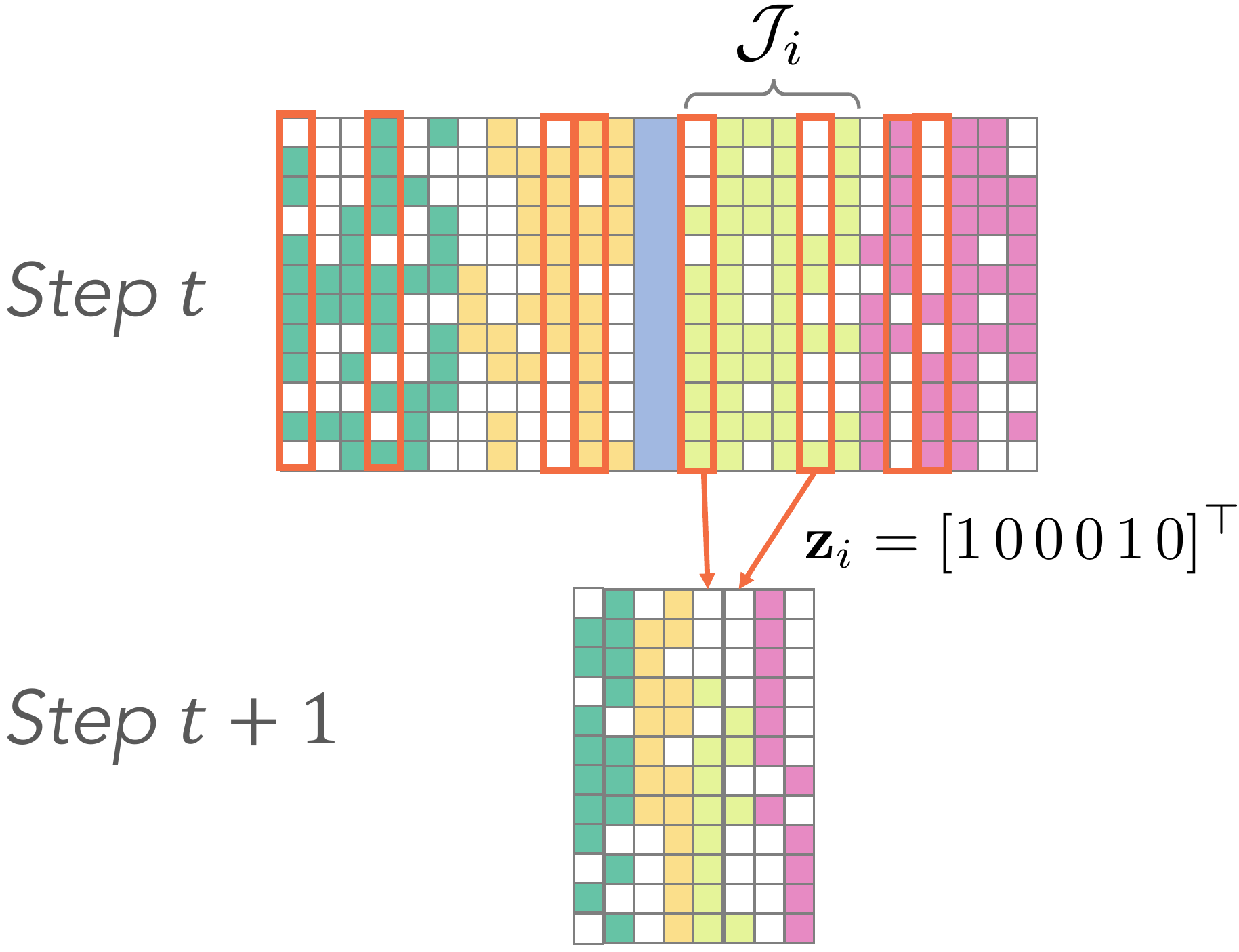}
    \caption{Illustration of the iterative pruning technique applied to $P$ in \ourmethoddec over consecutive iterations.
    At time $t$, $P$ is partitioned in different sub-problems of the same size (in the figure, 4 colour-coded problems with 6 models each) which are independently solved with \ourmethod.
    At time $t+1$, models belonging to the sub-solutions (highlighted in red in the figure) are retained; the others are discarded.}
    \label{fig:decomposed}
    \vspace{-0.75em}
\end{figure}

Our iterative pruning procedure is named \ourmethoddec and is detailed in Algorithm~\ref{algo:decomposition}.
Instead of directly solving the problem encoded by the preference matrix $P$, at each iteration we partition the columns of $P$ into multiple submatrices $P_{\mathcal{J}_{i}}$, where $\mathcal{J}_i$ indicates the set of selected column indices and $|\mathcal{J}_i|=s$. We choose the subproblem size $s$ so that the partitioned problems can be solved with high confidence by \ourmethod.
Each subproblem encoded by $P_{\mathcal{J}_{i}}$ is independently solved using \ourmethod (see Sec.~\ref{sec:mmf-qubo-formulation}), yielding an optimal binary vector $\mathbf{z}_i$ that represents the subset of selected models for the current subproblem. Note that models not selected in $\mathbf{z}_i$ are unlikely to be solutions to the original problem since, by construction, $\mathbf{z}_i$ %
maximizes the coverage on \textit{all points} with only the restricted subset of models $\mathcal{J}_i$. Then, after all subproblems have been solved, we prune $P$ by retaining only the columns that are selected in the corresponding optimal $\mathbf{z}_i$. 
This has the effect of significantly decreasing the dimensionality of the search space (see Fig.~\ref{fig:decomposed}).

We iteratively apply the above procedure until the number of remaining columns in $P$ falls below $s$, at which point the problem is directly and reliably solved using \ourmethod.

%% file: cready_sections/related.tex
\section{Related Work} 
\label{sec:related-work} 

The proposed approach is the first to address multi-model fitting using AQC. The most related work is \suter \cite{Doan_2022_CVPR} that leverages quantum computing as well for the scenario of single-model fitting.

\subsection{Multi-Model Fitting} 
\label{sec-related-mmf}

Robust fitting is a central problem in computer vision, and the case of multiple models counts several works that can be organized across two main directions, namely \emph{clustering-based}  and \emph{optimization-based} methods. 

Clustering-based methods cast multi-model fitting as a clustering task where %
points belonging to the same model must be grouped together. Several procedural algorithms have been presented leveraging different schemes: 
hierarchical clustering \cite{toldo2008robust,MagriFusiello14,MagriFusiello19,ZhaoZhangAl20,MagriLeveniAl21},  kernel fitting \cite{ChinWangAl09,ChinSuterAl10}, robust matrix factorization \cite{MagriFusiello17,TepperSapiro17}, biclustering \cite{TepperSapiro14,DenittoMagriAl16}, higher order clustering \cite{AgarwalJongowooAl05,Govindu05, JainGovindu13,ZassShashua05} and hypergraph partitioning \cite{PurkaitChinAl14, WangXiaoAl15,XiaoWangAl16,WangGupbao18,LinXiao19}.

Our approach falls instead in the category of optimization-based methods. Compared with procedural approaches based on clustering, methods based on the optimization of a precise objective function 
provide a quantitative criterion for evaluating the quality of the retrieved solution. 
In particular, our \ourmethod belongs to consensus-based approaches that generalize \textsc{RanSaC} by maximizing the consensus of the sought models.
Sequential \textsc{RanSaC} \cite{VincentLaganiere01}, Multi-\textsc{RanSaC}  \cite{zuliani2005multiransac} and \ransacov \cite{magrifusiello16} belong to this category as well. Other sophisticated techniques, that integrate consensus with additional priors, have also been proposed: Pearl \cite{IsackBoykov12}, 
Multi-X \cite{BarathMatas17} and  Prog-X\cite{BarathMatas19}. 

Our approach is related to \ransacov \cite{magrifusiello16} as we consider the same data structure, \textit{i.e.,} the preference matrix, and a similar optimization objective\footnote{The same objective is considered but constraints are different: \ransacov \cite{magrifusiello16} allows for overlapping models, resulting in an inequality constraint, see Problem \eqref{eq:set-cover}; instead, we consider disjoint models, using an equality constraint, see Problem \eqref{eq:disjoint-set-cover}. 
For a fair comparison, we slightly modify \ransacov, replacing the inequality with an equality constraint.}. 
However, there are substantial differences. 
\ransacov addresses the task via integer linear programming, as Magri and Fusiello \cite{magrifusiello16} use the \textsc{Matlab} function \texttt{intlinprog}, which employs linear programming relaxation followed by rounding the found solution, in a combination with some heuristics.
In case of failure it reduces to branch and bound. 
Our approach, instead, leverages quantum effects to optimize the objective directly in the space of qubits, where global optimality is expected with high probability after multiple anneals. 

\subsection{Quantum Computer Vision}
\label{sec:related_quantum}

Recently, many vision problems have been formulated for an AQC in QUBO forms, hence giving rise to the new field of Quantum Computer Vision (QCV).
AQC are especially promising for problems involving combinatorial optimization (see Sec.~\ref{sec:background}) such as 
graph matching \cite{SeelbachBenkner2020,SeelbachBenkner2021}, multi-image matching via permutation synchronization \cite{QuantumSync2021}, single-model fitting \cite{Doan_2022_CVPR}, multi-object tracking \cite{Zaech_2022_CVPR}, removing redundant boxes in object detection \cite{LiGhosh2020}, and motion segmentation \cite{Arrigoni2022}. 
Moreover, solving non-combinatorial problems 
on quantum hardware is of high general interest as well, since new formulations have unique properties, can be compact and would allow seamless integration with other quantum approaches in future. 
Thus, several methods address point set alignment  \cite{golyanik2020quantum, Meli_2022_CVPR} and approximate rotation  matrices by the exponential map with power series. 

Depending on the problem, QUBO formulation can be straightforward  \cite{LiGhosh2020,SeelbachBenkner2020,Zaech_2022_CVPR} or require multiple analytical steps 
bringing 
the initial objective to the quadratic form \cite{QuantumSync2021,Doan_2022_CVPR,Arrigoni2022}. 
In most cases, problem-specific constraints must be included in a QUBO as soft linear regularizers, 
as done in~\eqref{eq:qubo_soft}. 
This strategy is followed by the majority of works in the literature, including \cite{SeelbachBenkner2020,QuantumSync2021,Zaech_2022_CVPR,Arrigoni2022,Doan_2022_CVPR} and our approach.
Alternatively, it has been recently shown that it is possible to tackle linearly-constrained QUBO problems via the Frank-Wolfe algorithm \cite{YurtseverBirdalAl22}, which alleviates the need for hyper-parameter tuning, at the price of iteratively solving multiple QUBO problems and sub-linear convergence. 

All available approaches can be classified into single QUBO (``one sweep'') and iterative methods.
The first category prepares a single QUBO that is subsequently sampled on an AQC to obtain the final solution  \cite{LiGhosh2020,golyanik2020quantum,SeelbachBenkner2020, QuantumSync2021, Arrigoni2022}. These methods can solve comparably small problems \emph{on current hardware}: in the recent QuMoSeg \cite{Arrigoni2022}, the maximum problem size solved with high accuracy has 200 qubits, which correspond to five images with two motions having ten points each. 
In QuantumSync\cite{QuantumSync2021}, real experiments involve matching four keypoints in four images only. 
The second category of methods, instead, alternates between QUBO preparation on a CPU and QUBO sampling on an AQC, until convergence or a maximum number of iterations is reached \cite{SeelbachBenkner2021, Meli_2022_CVPR, Doan_2022_CVPR}. 
This is done either to overcome the limit on the maximum problem size solvable on current quantum hardware \cite{SeelbachBenkner2021, Meli_2022_CVPR} or because the QUBO represents only a small portion of the whole problem, hence requiring additional (classical) steps as in Q-Match \cite{SeelbachBenkner2021} and \suter  \cite{Doan_2022_CVPR}. Closely related to iterative methods are decompositional approaches (e.g., \cite{Zaech_2022_CVPR}) that partition the original objective into QUBO subproblems. 
This paper proposes both a single QUBO method (see  Sec.~\ref{sec:mmf-qubo-formulation}) and a decompositional, iterative pruning approach (see Sec.~\ref{sec:id-qmmf}). 

\textbf{Differences to \suter.} 
Most related to ours is a hybrid quantum-classical approach for robust fitting, \suter \cite{Doan_2022_CVPR}, that can be regarded as a pioneer in bringing robust fitting into a QUBO-admissible form, thus paving the way to the adoption of AQCs for such a class of hard combinatorial problems.
\suter and our approach have substantial differences. 
First, although being both based on linear programs, the two formulations do not share common steps: \suter considers the hypergraphs formalism, relying on multiple QUBOs in an iterative framework; our formulation, instead, is more compact as it involves a single QUBO in its one-sweep version. 
Second, \suter considers a \emph{single} model, whereas we consider the more general case of \emph{multiple} models, which is relevant in practical applications and more challenging in terms of the search space and the number of unknowns involved. 
Extending Doan \textit{et al.}~\cite{Doan_2022_CVPR} to MMF is not  straightforward (it is not clear so far whether it is possible). Finally, \suter is evaluated with an AQC on \emph{synthetic} data only, whereas we test our method on several {synthetic and} \emph{real} datasets on real quantum hardware, as shown in Sec.~\ref{sec:experimental-results}. Fig.~\ref{fig_suter} summarizes the core differences between \suter \cite{Doan_2022_CVPR} and our approach. 

{%
\textbf{Remark.} Finally, it is also worth mentioning the technique for robust fitting (single model) by Chin \textit{et  al.}~\cite{ChinSuterAl20} 
evaluated on a simulator and designed for a different type of quantum hardware than AQC, \textit{i.e.,} a gate-based quantum computer (GQC). 
GQCs operate differently from AQCs, \textit{i.e.,} they execute unitary transformations on quantumly-encoded data, and currently possess a much smaller number of qubits compared to AQCs \cite{Jurcevic2021}. 
We focus on methods that can run on real quantum hardware.
Unfortunately, modern AQCs cannot run algorithms designed for GQCs. 
}

%% file: cready_sections/experiments.tex
\section{Experiments} \label{sec:experimental-results}

In this section, we illustrate several experiments on synthetic and real datasets to evaluate the proposed approaches. 
The chosen metric is the \emph{misclassification error}, that is the percentage of wrongly classified points, as it is common practice in this field. Our comparisons mainly concentrate on the classical \ransacov \cite{magrifusiello16} and the hybrid quantum-classical \suter \cite{Doan_2022_CVPR}, which are the most related to our approaches, as explained in Sec.~\ref{sec:related-work}.
Quantum experiments run on the DWave Advantage System 4.1, a modern AQC by DWave built with the Pegasus topology \cite{Boothby2020}, comprising approximately 5000 physical qubits. Communication with the AQC is managed via the Ocean software stack \cite{OceanTools2021} and every experiment is executed with an annealing time of 20$\mu s$.

\smallskip
\textbf{Selection of $\lambda$.} In every experiment, we set $\lambda = 1.1$. 
This value is obtained with a two-step procedure: 
1) Initially, following common practices (see, \emph{e.g.}, \cite{QuantumSync2021,Arrigoni2022}), we perform an exhaustive grid search using \ourmethod to solve small-scale problems (encoded by at most $20$ qubits), from which we get an initial guess for $\lambda$ by looking at the \emph{optimal solution probability} \footnote{We refer to \emph{optimal solution probability} as the ratio between the anneals ended in a globally optimal state and the total number of anneals.}; 
2) Then, we evaluate the initial guess further, solving problems encoded by at most 100 qubits with \ourmethodsa, verifying whether or not the initial guess for $\lambda$ generalizes well to larger problems, too. 
In this second step, the reference metric is the misclassification error.

\begin{figure}[t]
    \centering
    \includegraphics[width=\columnwidth]{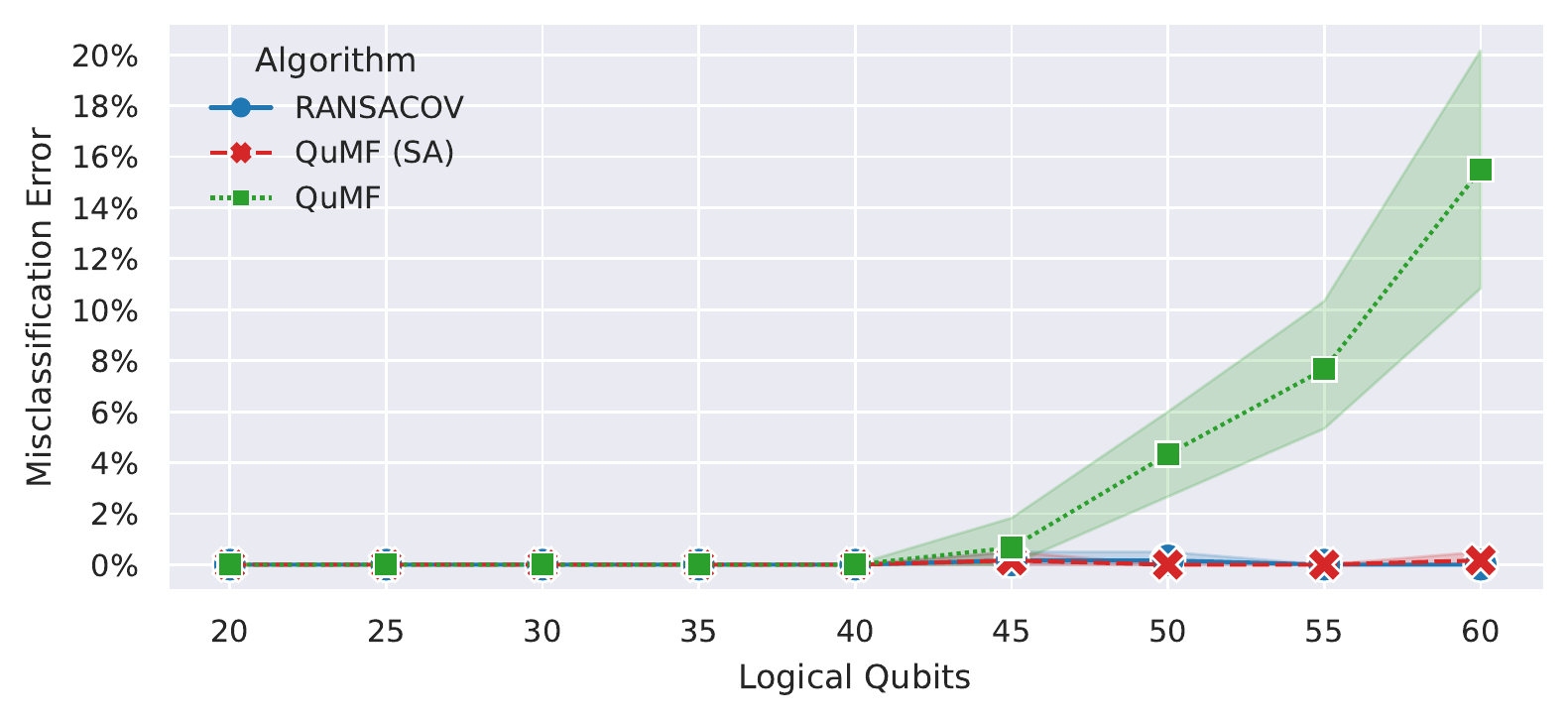}
    \vspace{-20pt}
    \caption{Misclassification Error for several methods on the \emph{Star5} dataset \cite{toldo2008robust}. The number of points $n$ is fixed to $n = 30$, the number of ground-truth structures is fixed to $k = 5$, and the number of sampled models $m$ is arranged on the x-axis. \vspace{-1em}
    }
    \label{fig:small_problems}
\end{figure}

\begin{figure}
    \centering
    \includegraphics[width=0.8\columnwidth]{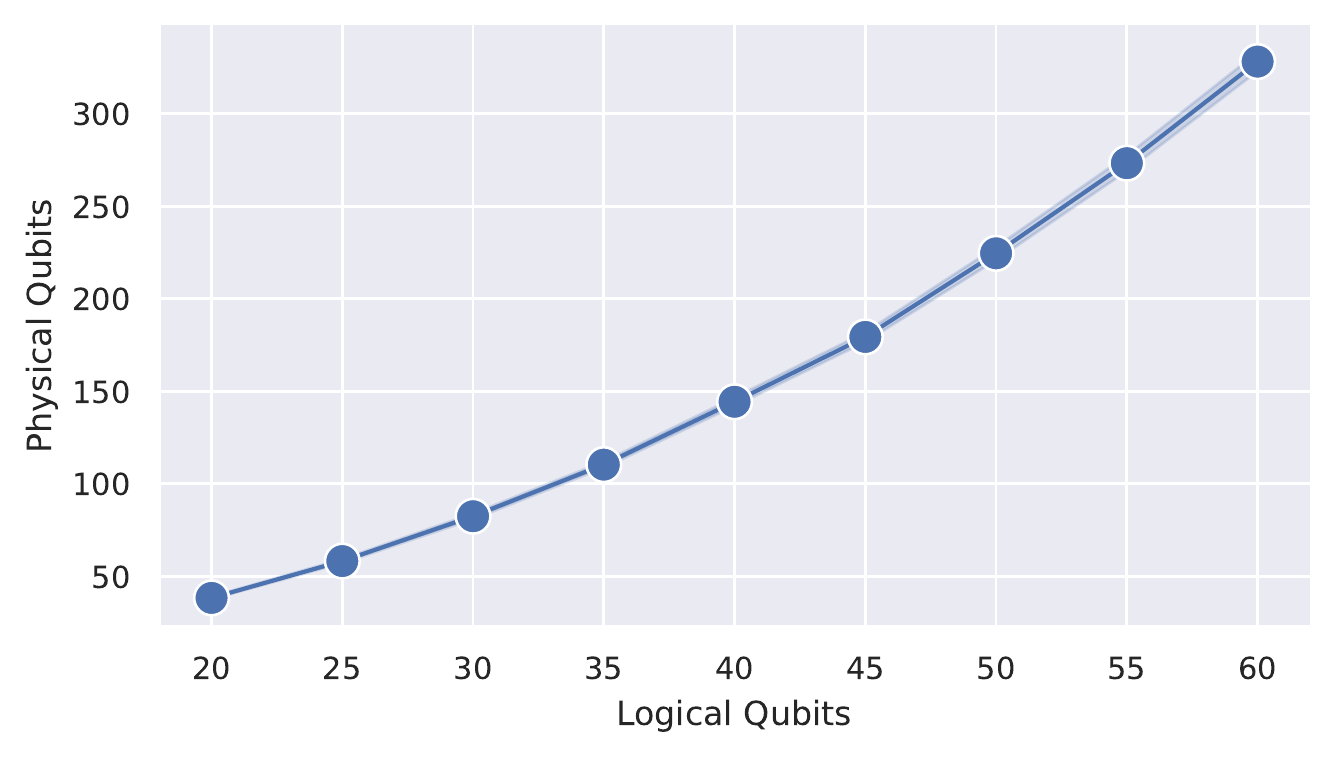}
    \vspace{-10pt}
    \caption{Relationship between \emph{physical qubits} and \emph{logical qubits} in embeddings produced with small-scale preference matrices from the \emph{Star5} dataset \cite{toldo2008robust}. \vspace{-1.5em}}
    \label{fig:phylog}
\end{figure}

\begin{figure}
    \centering
    \includegraphics[width=\columnwidth]{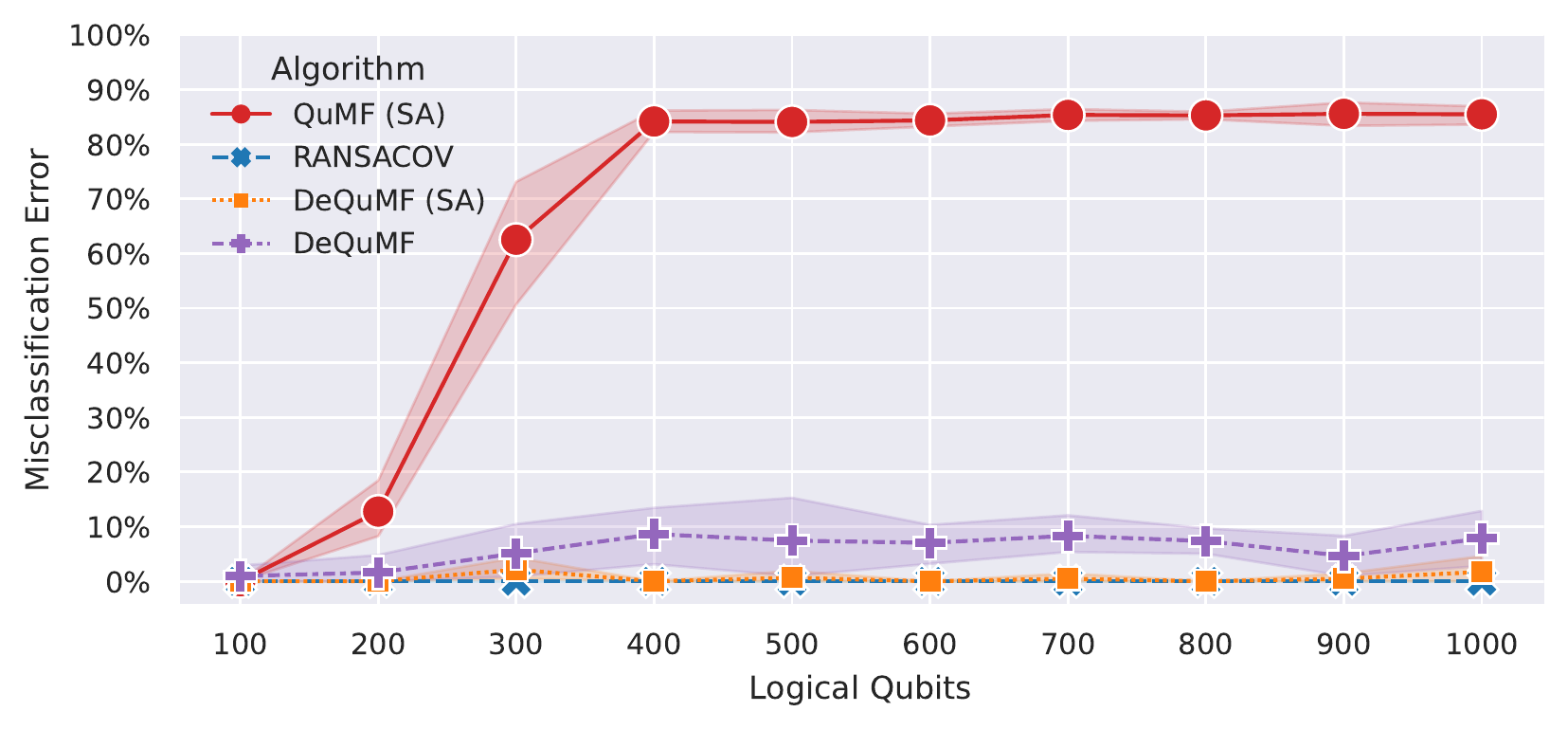}
    \vspace{-20pt}
    \caption{Misclassification Error for several methods on the \emph{Star5} dataset \cite{toldo2008robust}. The number of points $n$ is fixed to $n = 250$, the number of ground-truth structures is fixed to $k = 5$, and the number of sampled models $m$ (corresponding to the dimension of the search space) is arranged on the x-axis.
    \vspace{-2em}
    }
    \label{fig:star5_increasing_noise}
\end{figure}

\subsection{Synthetic Data}
\label{sec:results-synth}

We consider the synthetic \emph{Star5} dataset \cite{toldo2008robust}, that is used to fit five lines (depicting a star) given a set of 2D points. Outliers are removed to satisfy our assumptions.  

\smallskip
\textbf{Small-scale Data.}
In this experiment, we select a subset of $n=30$ points. 
The number of ground-truth models is fixed to $k = 5$, and additional $m-k$ excess models are sampled in order to study performances with increasing problem size. 
Note that this corresponds to including false models (i.e., columns) in the preference matrix, thereby representing noise.
For each configuration, 20 preference matrices are generated and averaged results are reported.
We consider our single-QUBO quantum approach \ourmethod and its CPU counterpart \ourmethodsa.
We also compare our approaches with \ransacov \cite{magrifusiello16}, whose code is available online\footnote{\url{https://fusiello.github.io/demo/cov/index.html}}, that can be viewed as a baseline implementing a sophisticated integer programming solver. 
All the methods receive as input the same preference matrix and optimize the same objective, as already observed in Sec.~\ref{sec-related-mmf}. 

Results are given in Fig.~\ref{fig:small_problems}, showing that \ourmethod can exactly solve problems up to 40 qubits, but it can not solve larger problems with high accuracy: this is an expected behaviour (in line with previous quantum papers, see Sec.~\ref{sec:related_quantum}), as \emph{current} quantum hardware is limited and far from maturity.
It is worth noting that our \ourmethodsa achieves excellent results (on par with \ransacov) for all the analyzed problem sizes: considering that \ourmethodsa uses the same QUBO as \ourmethod, we can thus view the former as an indication on the performances the latter can achieve in the future thanks to the continuous improvements in the quantum hardware.

\smallskip
\textbf{Physical vs Logical Qubits.}
We analyze here how the allocation of QPU resources (\emph{physical qubits}) grows with increasing problem size (\emph{logical qubits}). Similarly to our previous experiment, we use synthetic preference matrices generated from the \emph{Star5} dataset with increasing number of columns. However, we consider here all of the 250 points in the dataset in order to study denser cases. Our analysis is displayed in Fig.~\ref{fig:phylog}, showing that physical qubits grow super-linearly with respect to logical ones.
Such relationship makes it infeasible to allocate real world problems, as they typically require thousands of logical qubits. 
This finding, \emph{given the current state of the hardware}, strengthens the need for a decomposition approach.
Hence, we analyze our decomposition policy in the next paragraph.

\smallskip
\textbf{Large-scale Data.}
In this experiment, we consider up to 1000 models in the input preference matrix (also including the $k=5$ ground-truth models) as well as all of the $n=250$ points in the \emph{Star5} dataset. Since \ourmethod is not able to manage such large-scale data, it is not considered here. Instead, it is used as a sub-component of our decomposed approach \ourmethoddec: each sub-problem has a dimension that can be solved with high accuracy by \ourmethod, namely 20 qubits. %
As a reference, we also include \ourmethodsa and its decomposed version \ourmethodsadec as well as \ransacov \cite{magrifusiello16}.

Results are given in Fig.~\ref{fig:star5_increasing_noise}, showing that our \ourmethodsadec achieves nearly zero misclassification error in all the cases, on par with \ransacov. 
Average \ourmethoddec errors lie in the range $[0,10]\%$ for all problem sizes, showing the great effectiveness of the decomposition principle.
As for \ourmethodsa, this specific dataset represents a failure case: after inspecting the solution, we hypothesize that a possible reason is the sparsity pattern of the preference matrix, causing the algorithm to select more models than needed.

\begingroup
\renewcommand{\arraystretch}{1.25}
\begin{table*}[t]
    \caption{Misclassification Error [\%] on the \emph{Traffic3} and \emph{Traffic2} subsets of the Hopkins benchmark \cite{TronVidal07} 
    and on the York DB \cite{Denis2008ECCV}.
    \vspace{-1em}}
    \centering{\scriptsize
    \begin{tabular}{cccccccccccc}
         \toprule
         & & Multi-X \cite{BarathMatas17} & J-Linkage \cite{toldo2008robust} & T-Linkage \cite{MagriFusiello14} & RPA \cite{MagriFusiello17}
         & \ransacov \cite{magrifusiello16} & \ourmethodsa & \ourmethoddec & \ourmethodsadec \\
         \midrule
         \multirow{2}{6em}{\emph{Traffic3} \cite{TronVidal07}} & \emph{mean} & 0.32 & 1.58 & 0.48 & \textbf{0.19} & 0.35 & 5.14 & 8.74 & {0.55}\\
         & \emph{median} & \textbf{0} & 0.34 & 0.19 & \textbf{0} & 0.19 & 2.85 & 7.50 & 0.28\\
         \midrule
         \multirow{2}{6em}{\emph{Traffic2} \cite{TronVidal07}} & \emph{mean} & \textbf{0.09} & 1.75 & 1.31 & 0.14 & 0.54 & 6.04 & - & {0.10}\\
         & \emph{median} & \textbf{0} & \textbf{0} & \textbf{0} & \textbf{0} & \textbf{0} & 3.17 & - & \textbf{0}\\
         \midrule
         \multirow{2}{6em}{\emph{York} \cite{Denis2008ECCV}} &\emph{mean} & - & 2.85 & 1.44 & {1.08} & \textbf{0.19} & 12.29 & - & {0.74} \\
         & \emph{median} & - & 1.80 & \textbf{0} & \textbf{0} & \textbf{0} & 3.78 & - & \textbf{0} \\
    \bottomrule
    \end{tabular}
    \vspace{-1em}
    }
    \label{tab:hopkins}
\end{table*}
\endgroup

\subsection{Real Data}
\label{sec:results-real}

In this section, we test our framework on several real world datasets, each associated to a different task. Since these are large-scale problems, we do not consider \ourmethod but its CPU variant \ourmethodsa. %

\smallskip
\textbf{AdelaideRMF dataset \cite{wong2011dynamic}.}
From this benchmark, we consider the 15 sequences involving multiple fundamental matrices (which, in turn, can be used to perform motion segmentation in two images) and we remove outliers beforehand. All the analyzed methods are given the same preference matrices as input. Results are given in Tab.~\ref{tab:adel_mmf}. 
Both \ourmethodsa and \ourmethodsadec significantly outperform \ransacov \cite{magrifusiello16} in this scenario. 
In particular, \ourmethodsadec performs better than its quantum version \ourmethoddec, in agreement with our previous experiments.

\begingroup
\renewcommand{\arraystretch}{1.25}
\begin{table}[t]
    \centering{\footnotesize
    \resizebox{1\columnwidth}{!}
    {
    \begin{tabular}{ccccc}
         \toprule
         & \ransacov \cite{magrifusiello16} & \ourmethodsa & \ourmethoddec & \ourmethodsadec \\
          \midrule
         \emph{mean} & 9.79 & 3.85 & 16.22 & \textbf{0.77}\\
         \emph{median} & 7.97 & 3.54 & 11.0 & \textbf{0.18}\\
         \bottomrule
    \end{tabular} 
    }
    }
    \caption{Misclassification Error [\%] for several methods on the 15 \textbf{multi-model} sequences of the AdelaideRMF dataset \cite{wong2011dynamic}. \vspace{-2em}}
    \label{tab:adel_mmf}
\end{table}
\endgroup

\smallskip
\textbf{Hopkins benchmark \cite{TronVidal07}.}
From this dataset, we consider the \emph{Traffic2} and \emph{Traffic3} sequences, which comprise 31 and 7 scenes, respectively. Here, the task is to fit multiple subspaces to point trajectories in the context of video motion segmentation. This benchmark provides outlier-free data, hence no pre-processing is needed. Results are given in Tab.~\ref{tab:hopkins}: in addition to \ransacov \cite{magrifusiello16}, other multi-model fitting approaches have been considered, whose results are taken from the respective papers. All the variants of our approach\footnote{We are not able to run \ourmethod on \emph{Traffic2} due to the limited amount of QPU time available in our monthly subscription combined with the large dimension of the dataset, hence results are not reported.} perform reasonably well, with \ourmethodsadec being on par with the state of the art.%

\smallskip
\textbf{York Urban Line Segment Database \cite{Denis2008ECCV}.}
This dataset contains 102 images used for vanishing point detection from multiple line segments without outliers.
Results are presented in Table \ref{tab:hopkins}, %
following the same rationales as per the Hopkins benchmark. %
Such results are in line with our previous experiments: in particular, \ourmethodsadec exhibits competitive performances with the state of the art.

\subsection{Comparison with Quantum State-of-the-art}
\label{sec:quantum_comparison}
Our method is the first quantum solution to \emph{multi-model} fitting, hence there are no quantum competitors in the literature.
The most similar work to ours is \suter \cite{Doan_2022_CVPR}, whose code is available online\footnote{\url{https://github.com/dadung/HQC-robust-fitting}}, which addresses \emph{single-model} fitting in the presence of outliers (see Sec.~\ref{sec:related_quantum}).
\suter has strong theoretical justifications, and the proposed algorithm terminates with either the globally optimal solution or a feasible solution paired with its \emph{error bound}.
Neither \ourmethod or \ourmethoddec have this property.
On the other hand, our framework is general and can be used for a single model as well, allowing for an experimental comparison with \suter \cite{Doan_2022_CVPR}. 
Specifically, we exploit the information that only a single model is present in the data and we select, from the solution returned by set-cover, only the model with the maximum consensus set. All the other models in the solution are treated as outliers.

We focus on fundamental matrix estimation and use the 4 sequences of the AdelaideRMF dataset \cite{wong2011dynamic}  characterized by the presence of a single model. 
Since this is a large-scale dataset, we omit \ourmethod and consider \ourmethodsa, \ourmethoddec and \ourmethodsadec, as previously done.
We analyze how performances vary with different outlier ratios: we consider $10\%$ and $20\%$ of outliers, in addition to the evaluation on the original sequences, where no fixed outlier ratio is imposed.
Results are summarized in Tab.~\ref{tab:adelrmf_single}.

\begin{figure}
    \centering
    \includegraphics[width=0.48\columnwidth]{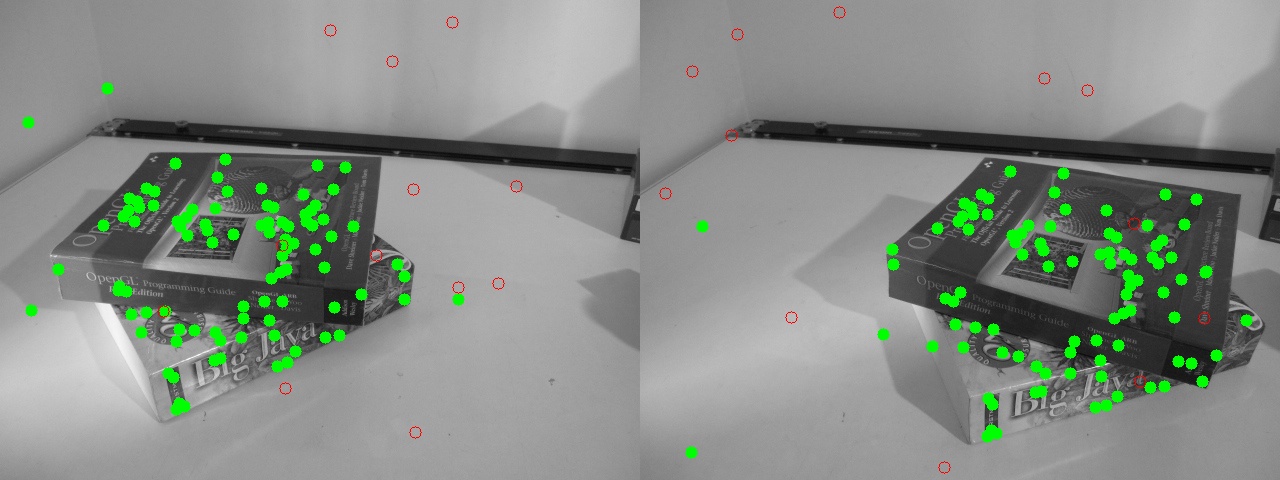}
     \includegraphics[width=0.48\columnwidth]{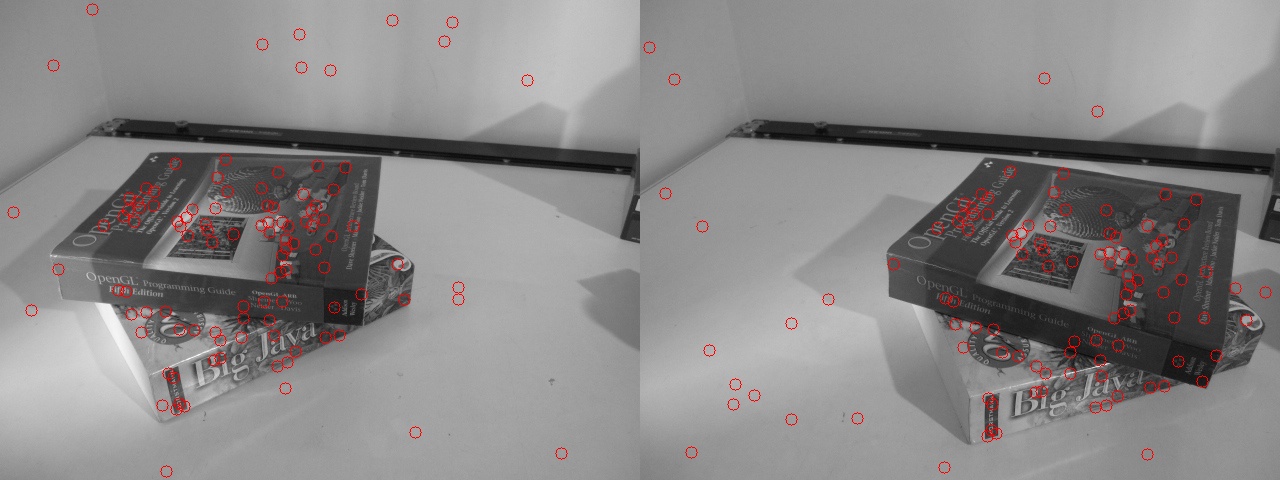} 
    \ourmethodsadec \quad\quad\quad\quad  \suter \cite{Doan_2022_CVPR}  \\ 
    \caption{
    A sample of the worst results of the considered quantum methods on the \emph{book} image pair from the AdealideRMF  \cite{wong2011dynamic} with $20\%$ outliers. Outliers and inliers are shown in red and green, respectively. On the right, \suter \cite{Doan_2022_CVPR} classifies all the points as outliers, whereas our method achieves $8.49\%$ error.}
    \vspace{-1.75em}
    \label{fig:qualitative}
\end{figure}

\suter shows remarkable behaviour in controlled scenarios with low outlier ratios; however, it is worth noting that our method is more robust to higher outlier ratios, although not explicitly designed for single-model fitting.
See also Fig.~\ref{fig:qualitative} for qualitative results of worst-case scenarios.

\begin{table}
    \centering
    \resizebox{0.75\columnwidth}{!}
    {\small
    \begin{tabular}{ccccccc} 
    \toprule
    & \multicolumn{3}{c}{Outlier ratio} \\
   \cmidrule(lr){2-4} Algorithm & 10\% & 20\% & Full sequences \\
    \midrule
    \ourmethodsa & 7.22 & 11.34 & 13.23 \\
    \ourmethoddec & \textbf{2.41} & 10.53 & 16.17\\
    \ourmethodsadec & 6.26 & \textbf{8.28} & \textbf{10.83} \\
    \suter \cite{Doan_2022_CVPR} & 3.71 & 37.0 & 45.84 \\
    \bottomrule
    \end{tabular}
}
    \caption{Misclassification Error [\%] for quantum methods on the \textbf{single-model} sequences of the AdelaideRMF dataset \cite{wong2011dynamic}. \vspace{-1em}}
    \label{tab:adelrmf_single}
\end{table}

%% file: cready_sections/limitations.tex
\section{Limitations}
\label{sec:limitations}
Differently than current quantum methods for single-model robust fitting \cite{Doan_2022_CVPR}, our approach does not terminate with a solution paired with its \emph{error bound}.  
Additionally, the decomposition policy at the core of \ourmethoddec does not provide theoretical guarantees of global optimality, as globally optimal models may be discarded with locally optimal selection performed by \ourmethod.
We believe it is worth trying to investigate this research question in future work.

%% file: cready_sections/conclusion.tex
\section{Conclusion} 
\label{sec:conclusion}
We brought a new, very important problem into AQC-admissible form.
Our method shows promising results on a variety of datasets for MMF and even surpasses the quantum SotA on single-model fitting, although not explicitly designed for such task.
We believe it offers many avenues for future research.

{\small
\paragraph{Acknowledgements.} 
This paper is supported by FAIR (Future Artificial Intelligence Research) project, funded by the NextGenerationEU program within the PNRR-PE-AI scheme (M4C2, Investment 1.3, Line on Artificial Intelligence).
This work was partially supported by the PRIN project LEGO-AI (Prot. 2020TA3K9N) and it was carried out in the Vision and Learning joint laboratory of FBK and UNITN.
}

%% file: cready_sections/suppmat.tex
\clearpage
\appendix
\begin{center}
\textbf{\Large Supplementary Material} 
\end{center} 
\section{Parameters and Implementation Details} 
This section highlights the parameter settings of the experiments presented in Sec.~\ref{sec:experimental-results} of the main paper. 

\begin{figure}[b!]
    \centering
    
    \begin{subfigure}{\columnwidth}
    \centering
    \includegraphics[width=\columnwidth]{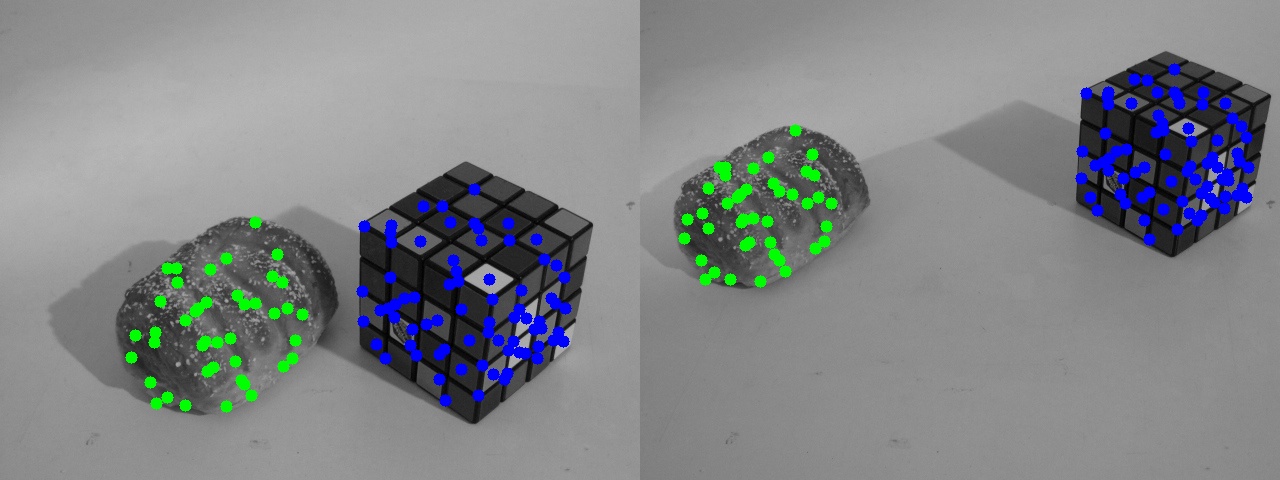}
    \caption{\ourmethoddec outcome, misclassification error = 1.7\%.}
    \label{fig:adel_qa_best}
    \end{subfigure}
    
    \begin{subfigure}{\columnwidth}
    \centering
    \includegraphics[width=\columnwidth]{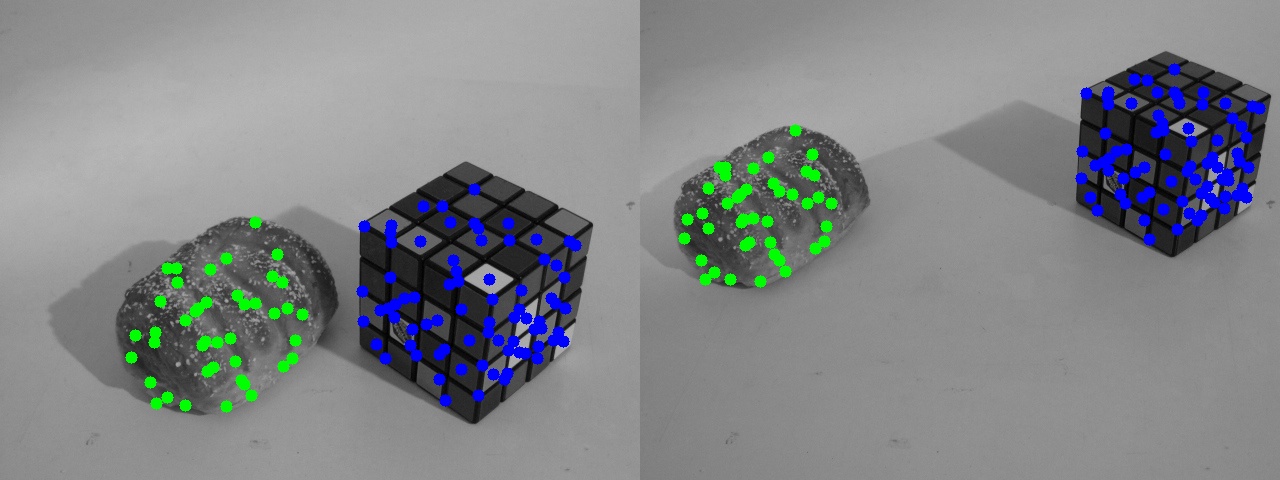}
    \caption{\ourmethodsadec outcome, misclassification error = 0\%.}
    \label{fig:adel_sa_best}
    \end{subfigure}
    
    \begin{subfigure}{\columnwidth}
    \centering
    \includegraphics[width=\columnwidth]{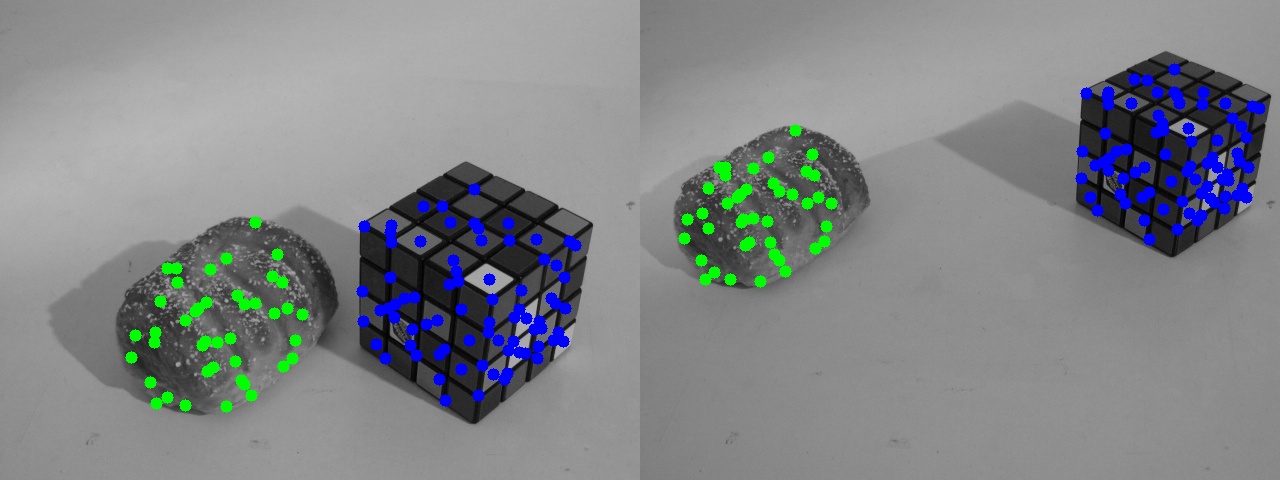}
    \caption{Ground-truth segmentation.}
    \label{fig:adel_best_gt}
    \end{subfigure}
    
    \caption{A sample of the \emph{best} case for our approach on the \emph{breadcube} sequence of the \emph{AdelaideRMF} dataset \cite{wong2011dynamic}.} 
    \label{fig:adel_qualitatives_best}
\end{figure}

\begin{figure}[b!]
    \centering
    
    \begin{subfigure}{\columnwidth}
    \centering
    \includegraphics[width=\columnwidth]{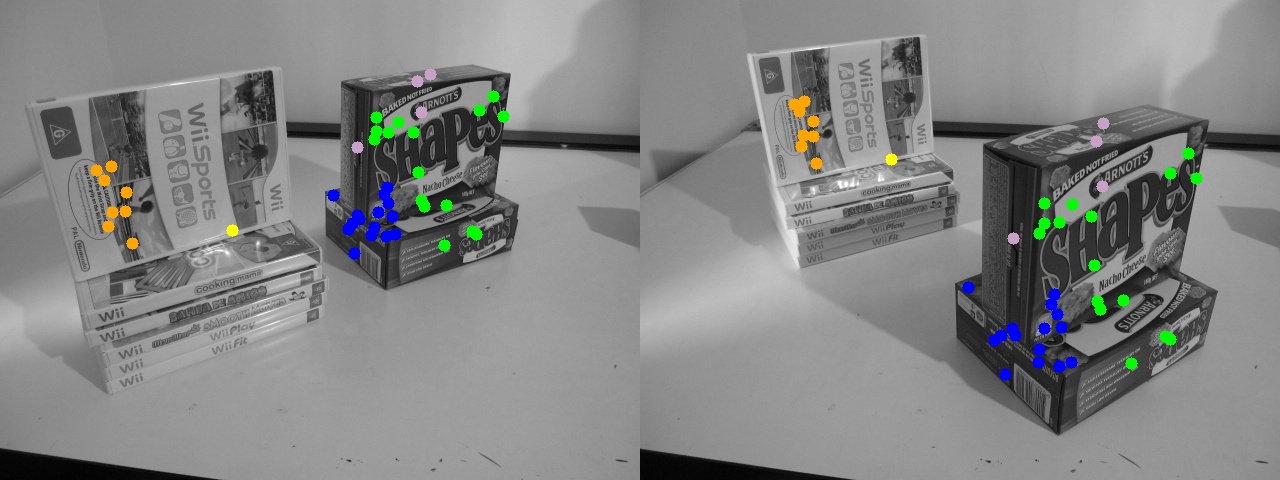}
    \caption{\ourmethoddec outcome, misclassification error = 47.9\%.}
    \label{fig:adel_qa_worst}
    \end{subfigure}
    
    \begin{subfigure}{\columnwidth}
    \centering
    \includegraphics[width=\columnwidth]{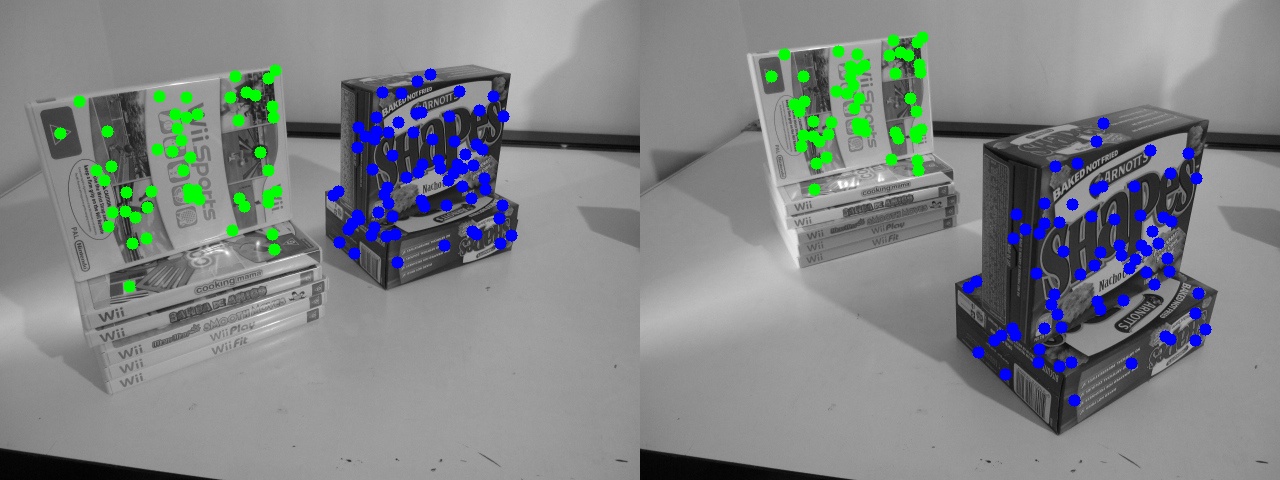}
    \caption{\ourmethodsadec outcome, misclassification error = 1.7\%.}
    \label{fig:adel_sa_worst}
    \end{subfigure}
    
    \begin{subfigure}{\columnwidth}
    \centering
    \includegraphics[width=\columnwidth]{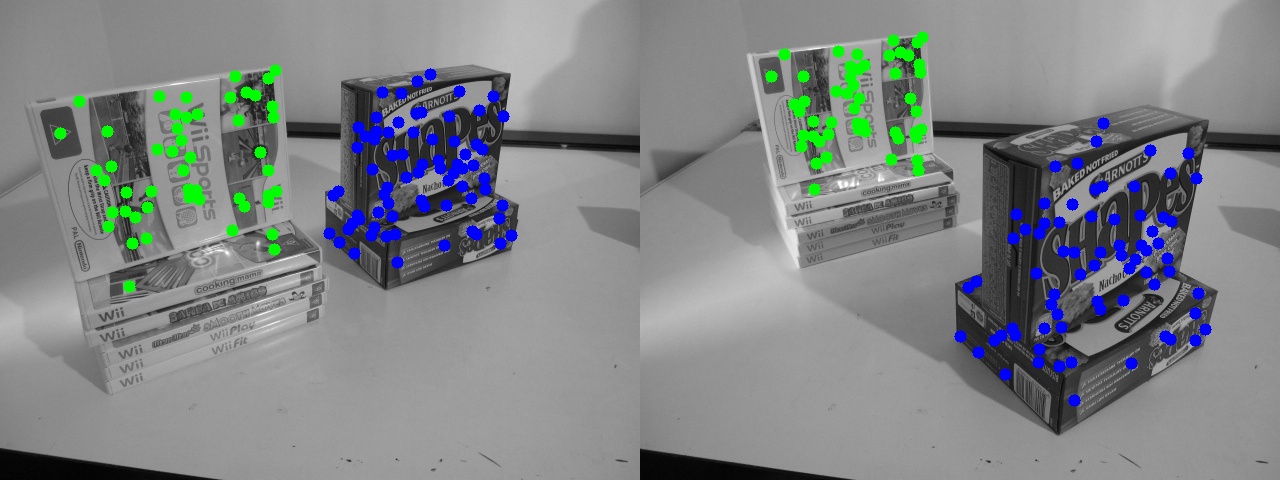}
    \caption{Ground-truth segmentation.}
    \label{fig:adel_worst_gt}
    \end{subfigure}
    
    \caption{A sample of the \emph{worst} case for our approach on the \emph{gamebiscuit} sequence of the \emph{AdelaideRMF} dataset \cite{wong2011dynamic}.} 
    \label{fig:adel_qualitatives_worst}
\end{figure}

\paragraph{The \emph{chain strength} parameter.}
Following previous work (e.g., \cite{QuantumSync2021,Arrigoni2022}), we employ the \emph{maximum chain length criterion} for \emph{all} our experiments: 
given a logical graph (i.e., a QUBO problem to be solved), we first minor-embed it to the physical one. Once the final embedding is found, the length $l$ of the longest qubit chain is computed. The chain strength parameter is then set to $l$ plus a small offset, \textit{i.e.,} $0.5$ in our experiments. 

\paragraph{Number of anneals.}
We set the number of anneals as follows:
algorithms solving QUBOs built with the entire preference matrix ("one-sweep" methods \ourmethod and \ourmethodsa) are executed with 5000 and 100 anneals, respectively.
Algorithms employing our iterative
decomposition principle %
are configured differently:
\begin{enumerate}
    \item \ourmethoddec executes 2500 anneals for each subproblem, with subproblems containing 20 models each;
    \item \ourmethodsadec runs with 100 anneals for each subproblem, with subproblems having 40 models.
\end{enumerate} 
In Sec.~\ref{sec:quantum_comparison}, when evaluating \ourmethoddec and \suter \cite{Doan_2022_CVPR}, to provide a fair comparison we fix the amount of available resources per-routine.
In this setup, we configured all the optimization routines (SAs/QAs) with 100 anneals, regardless of the algorithm.

\paragraph{Synthetic experiments.}
To collect statistics, we execute each CPU-based algorithm ten times and report average results.
Due to limited QPU availability in our subscription, \ourmethoddec is run 5 times.
In Figs.~\ref{fig:small_problems}-\ref{fig:star5_increasing_noise}, the 95\% confidence intervals are displayed around data points.
To provide a comparison on an equal basis, all the methods have been tested on the \emph{same} preference matrices. 
In addition, for this test we used a modified version of \textsc{RanSaCov} \cite{magrifusiello16}, which enforces disjointedness among the sets of the retrieved solution, as done in the (soft) constraint of \ourmethod.

\paragraph{Real experiments.} In the context of multi-model experiments of Sec.~\ref{sec:experimental-results} of the main paper (Tables~\ref{tab:hopkins}-\ref{tab:adel_mmf}), for a fair comparison, the  inlier threshold $\epsilon$ has been tuned per sequence as in \cite{magrifusiello16}. 
Preference matrices are instantiated  with a fixed points-to-models ratio, \textit{i.e.,} the number of models $m$ is always $m=\sigma n$. 
In our experiments, we select $\sigma=6$ for \emph{AdelaideRMF} \cite{wong2011dynamic} and the \emph{Hopkins} \cite{TronVidal07} benchmarks, $\sigma=10$ for the \emph{York} dataset \cite{Denis2008ECCV}.
Recall that the number of models is equal to the number of logical qubits, which essentially determines the problem complexity.
Additional insights on the characteristics of the involved datasets are provided in Tab.~\ref{tab:datasets}. 
Following synthetic experiments, CPU-based algorithms are run ten times and averaged results are reported; \ourmethoddec, when available, is executed once.

\begin{table}[t]
    \centering
    \resizebox{1\columnwidth}{!}
    {
    \begin{tabular}{cccccc}
         & & \emph{AdelaideRMF} \cite{wong2011dynamic} & \emph{Traffic2} \cite{TronVidal07} & \emph{Traffic3} \cite{TronVidal07} & \emph{York} \cite{Denis2008ECCV} \\
         \toprule 
         \multirow{3}{1.25em}{$n$} & \emph{mean} & 160 & 241 & 332 & 119\\
         & \emph{min} & 105 & 41 & 123 & 25 \\
         & \emph{max} & 239 & 511 & 548 & 627 \\
         \midrule
         \multirow{3}{1.25em}{$m$} & \emph{mean} & 960 & 1446 & 1994 & 1188\\
         & \emph{min} & 630 & 246 & 738 & 250 \\
         & \emph{max} & 1434 & 3066 & 3288 & 6270 \\
         \midrule
         \multirow{2}{1.25em}{$k$} %
         & \emph{min} & 2 & 2 & 3 & 2 \\
         & \emph{max} & 4 & 2 & 3 & 3 \\
         \midrule
         $\sigma$ & \emph{const} & 6 & 6 & 6 & 10\\
         \bottomrule
    \end{tabular}
    }
    \caption{Details of each real dataset used in the experiments of Sec.~\ref{sec:experimental-results} in the main paper: $n$ refers to the number of points, $k$ to the number of ground-truth structures, $m$ is the number of models computed as $m=\sigma n$.}
    \label{tab:datasets}
\end{table}

\paragraph{Runtimes for real experiments.}
Runtimes of the experiments executed on real world datasets are reported in Tab.~\ref{tab:runtimes} to provide completeness to our experimental evaluation.
However, these numbers may report a distorted comparison because of:
\begin{enumerate}
    \item \emph{Implementation differences}, as \ransacov  \cite{magrifusiello16} is MATLAB-based, whilst \suter \cite{Doan_2022_CVPR} and all of our methods are Python-based;
    \item \emph{Hardware differences}, highlighted in the caption of Tab. \ref{tab:runtimes};
    \item \emph{Overheads} related to AQCs; specifically, we are including \emph{network communication times} since we must access a shared AQC we do not have in loco, and \emph{resource allocation times} due to other users that would not be present with an in-house machine.
\end{enumerate} 
These technicalities currently make the comparison \emph{unfair}.
Additionally, we point out that the annealing time ($20\mu s$) is independent on the problem size.
Hence, in anticipation of stable AQCs, we believe our quantum approach is a prominent research direction in terms of runtime, too.

\begin{table}[t!]
    \centering
    \begin{adjustbox}{max width=\columnwidth}
    {
    \begin{tabular}{lccccc}
        \toprule
         &  {\emph{AdelaideRMF-S}} & {\emph{AdelaideRMF-M}} & {\emph{Traffic2}} & {\emph{Traffic3}} & {\emph{York}} \\
         \midrule
         $^{\dagger}$\ransacov \cite{magrifusiello16} & - & \textbf{0.48} & \textbf{1.46 } & \textbf{2.14}  & \textbf{0.24} \\
         $^{\dagger}$\ourmethodsadec (ours) & \textbf{0.99} & 0.77 & 2.54 & 3.65 & 1.93 \\
         $^{\ddagger}$\ourmethodsa (ours) & 23.92 & 10.72 & 53.05 & 76.46 & 102.38 \\
         $^{\ddagger}$\ourmethoddec (ours) & 89.62 & 116.71 & - & 376.09 & - \\
         $^{\ddagger}$\suter \cite{Doan_2022_CVPR} & 2349.27 & - & - & - & - \\
         \bottomrule
    \end{tabular}
    }
    \end{adjustbox}
    \caption{Mean runtimes [$s$]. \emph{AdelaideRMF-S} is single-model, with \emph{M} is multi-model. ``$^{\dagger}$'': run on an Intel-i7-8575U; ``$^{\ddagger}$'': run on an Intel-i9-7900X. Please zoom in.} 
    \label{tab:runtimes} 
\end{table}

\paragraph{Comparison with quantum state of the art.}
Concerning Tab.~\ref{tab:adelrmf_single}, we fix the inlier threshold as $\epsilon = 0.045$ for all the sequences for both our method and \suter \cite{Doan_2022_CVPR}.
Apart from $\epsilon$ (that essentially defines a bound on the distance of inlier points to models), when evaluating \suter, we use the source code provided by the authors and the default parameter settings therein. 

\paragraph{Additional insights on \ourmethod.} Recall that the quadratic term in our QUBO formulation is determined by the product $P^{\mathsf{T}}P$, with $P$ being the preference matrix. Hence, from an interpretative perspective, the quadratic term between models $i$ and $j$ (which we denote by $q_{ij}$ from now on) is equal to the number of points the two have in common. By leveraging this interpretation, a useful observation can be made for a very peculiar situation, namely the case where, for a fixed $i$, we have $q_{ij} = 0 \  \forall j \neq i$. This means that model $i$ does not share any point with the remaining models in $P$: In such a situation, we can conclude that model $i$ is an essential component of the final solution $\mathbf{z}$, as it covers {at least} one point that is not explained by any other model. In other terms, $\mathbf{z}_i = 1$ must be true at the global optimum. This simple observation thus allows to prune the search space and proceed with embedding a \emph{reduced} (i.e., smaller) QUBO. 
On the contrary, if the full logical graph was considered, then the minor embedding procedure would avoid mapping model $i$ to the QPU; this would result in unmapped logical qubits, which is undesirable in practice. 

\section{Qualitative Results}
\begin{figure}
    \centering
    
    \begin{subfigure}{\columnwidth}
        \centering
        \includegraphics[width=\columnwidth]{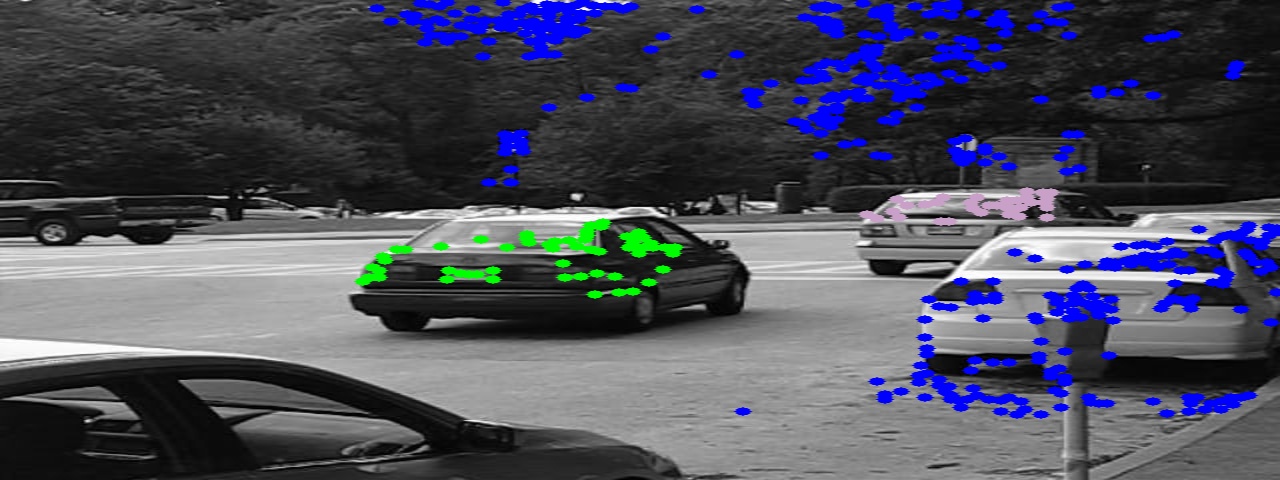}
        \caption{\ourmethoddec outcome, misclassification error = 0.2\%.}
        \label{fig:hopkins_qa_best}
    \end{subfigure}

    \begin{subfigure}{\columnwidth}
        \centering
        \includegraphics[width=\columnwidth]{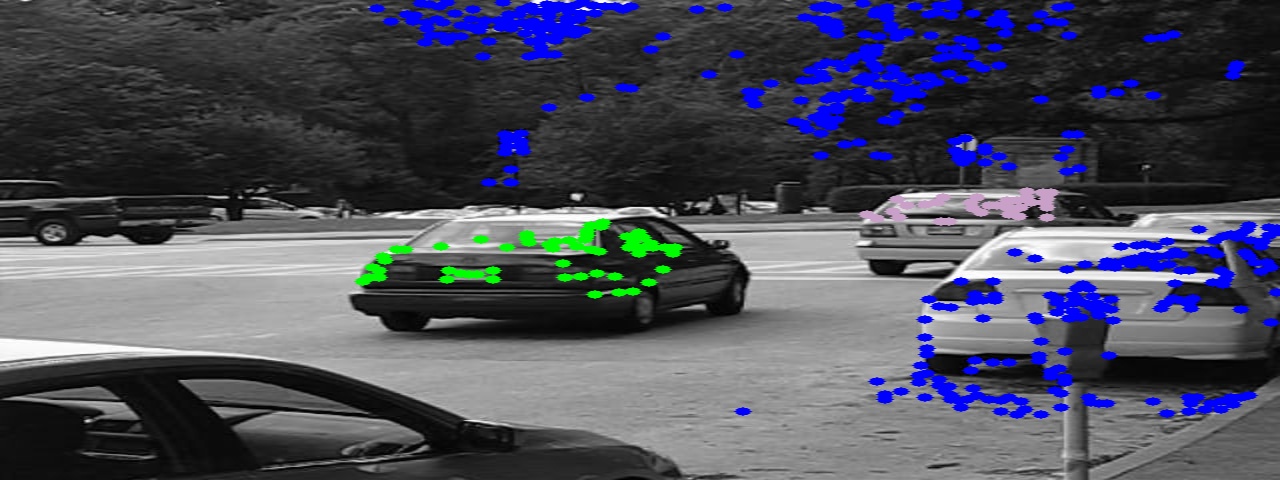}
        \caption{\ourmethodsadec outcome, misclassification error = 0\%.}
        \label{fig:hopkins_sa_best}
    \end{subfigure}
        
    \begin{subfigure}{\columnwidth}
        \centering
        \includegraphics[width=\columnwidth]{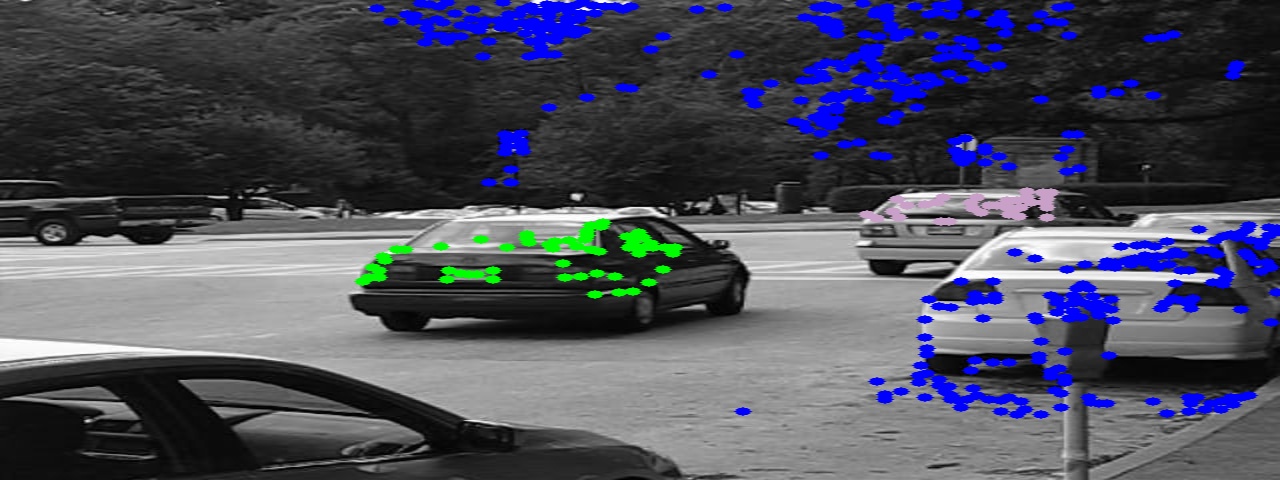}
        \caption{Ground-truth segmentation.}
        \label{fig:hopkins_gt}
    \end{subfigure}

    \caption{A sample case for our approach on the \emph{cars2B} sequence of the \emph{Hopkins} benchmark \cite{TronVidal07}. Point membership is colour-coded and ground-truth segmentation is also reported.} 
    \label{fig:hopkins_qualitatives_best}
\end{figure}

In this section, we provide qualitative results to visually interpret the behaviour of the proposed approaches. 
Results are given in Figs.~\ref{fig:adel_qualitatives_best}, \ref{fig:adel_qualitatives_worst} and \ref{fig:hopkins_qualitatives_best}.
In particular, a failure case is reported in Fig.~\ref{fig:adel_qa_worst} where \ourmethoddec exhibits pseudo-random behaviour. 
Apart from that, in general, both variants of our approach---\ourmethoddec and \ourmethodsadec---return accurate results, confirming the outcome of the quantitative analysis reported in the main paper.

\section{Minor Embeddings} 
\begin{figure}[h!]
    \centering
    \begin{subfigure}{\columnwidth}
    \centering
    \includegraphics[width=\columnwidth]{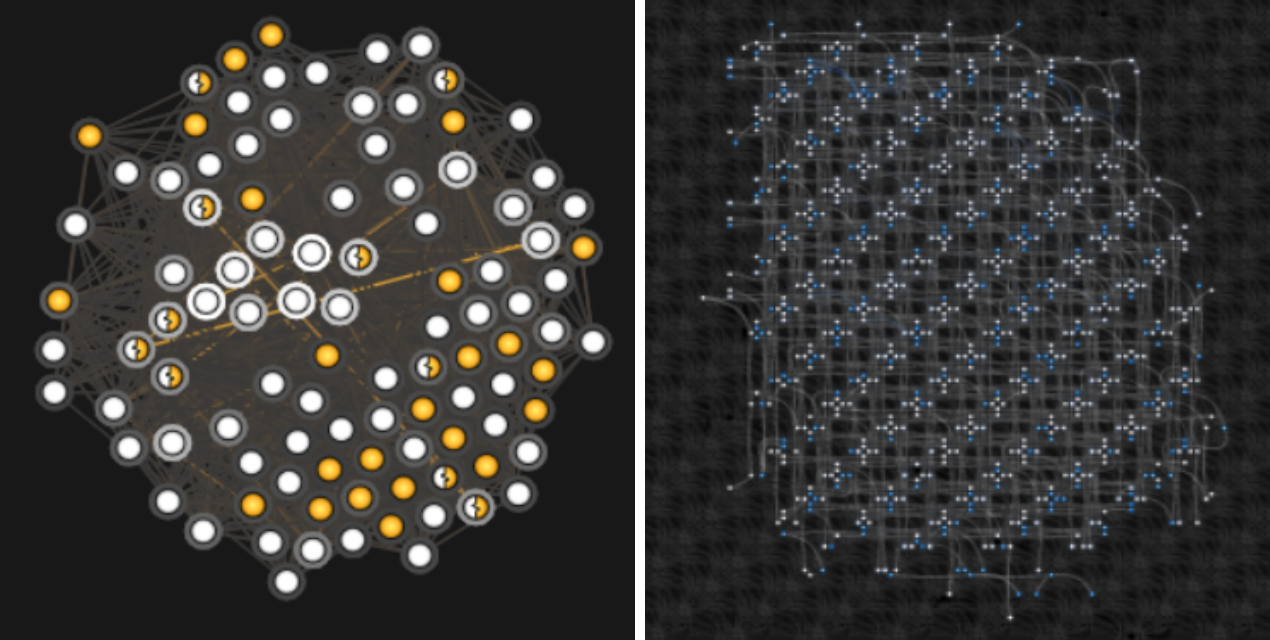}
    \caption{Visualization of a full QUBO from \emph{Star5} \cite{toldo2008robust}. The 100 nodes composing the logical graph (left) are mapped to 869 physical qubits (right).}
    \label{fig:fullstar5embed}
    \end{subfigure}

    \begin{subfigure}{\columnwidth}
    \centering
    \includegraphics[width=\columnwidth]{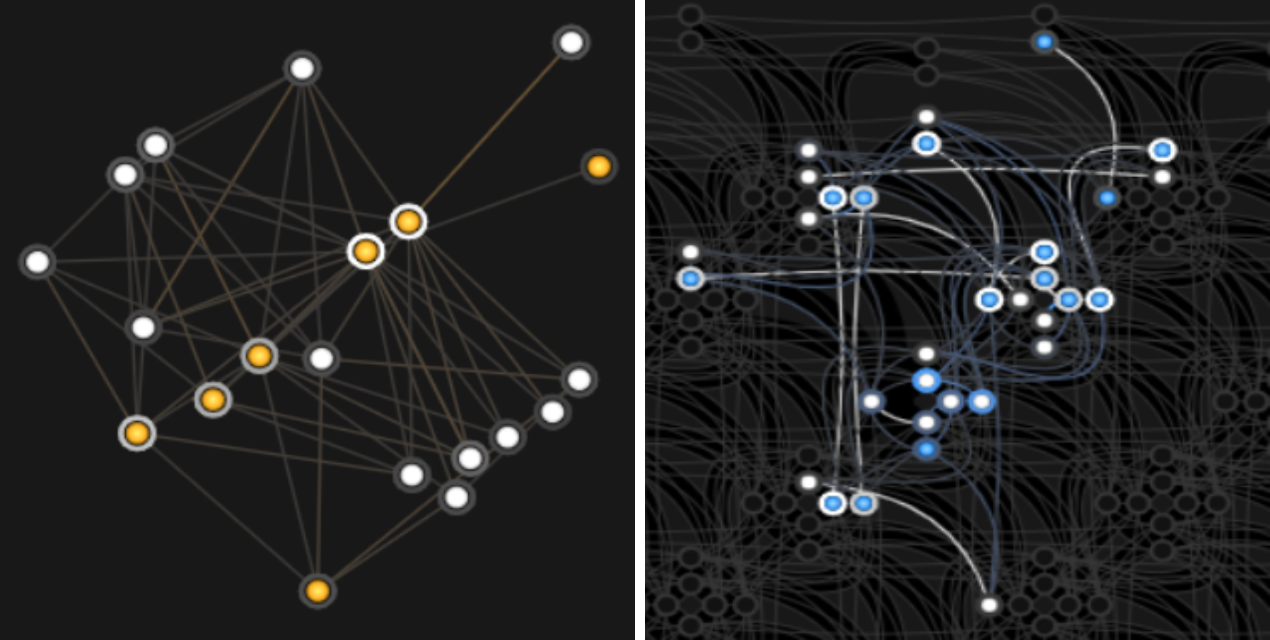}
    \caption{Visualization of a subproblem extracted from \emph{Star5} \cite{toldo2008robust}. The 20 nodes composing the logical graph (left) are mapped to 32 physical qubits (right).}
    \label{fig:substar5embed}
    \end{subfigure}
    
    \caption{
    Visual representation of logical and physical graphs when working with the \emph{Star5} dataset \cite{toldo2008robust}. Sub-figure (a) refers to a preference matrix with $n=250$ points and $m=100$ models while sub-figure (b) considers a small portion of such preference matrix, corresponding to a sub-problem with $m=20$ models (as tackled by our decomposed approach).
    } 
    \label{fig:star5embed} 
\end{figure}

Finally, we visually analyze the concepts of logical and physical graphs.
Specifically, Fig.~\ref{fig:star5embed} reports both the logical graph and its mapping to the QPU (obtained as a result of a minor embedding procedure) on exemplary problems with 20 and 100 logical qubits, respectively. 
Fig.~\ref{fig:star5embed} clearly shows that the problem with 100 logical qubits requires a significantly larger amount of physical qubits, as already observed in the main paper. \newpage

